\documentclass[a4paper, 11pt]{article} 
\usepackage[utf8]{inputenc}
\usepackage{geometry}
\usepackage{times} 
\usepackage[runin]{abstract}

\usepackage{paralist}

\usepackage{sectsty}
\allsectionsfont{\sffamily} 
\makeatletter
\def\@seccntformat#1{\csname the#1\endcsname.\quad}
\makeatother

\usepackage{titlesec}
\titleformat{\subsubsection}[runin]
{\normalfont\sffamily\bfseries}{\thesubsubsection}{1em}{}

\usepackage[colorlinks]{hyperref}
\AtBeginDocument{\hypersetup{citecolor=black, linkcolor=black, urlcolor=blue}}

\usepackage{graphicx} 

\usepackage[square, numbers]{natbib}
\usepackage[font=small,labelfont=bf]{caption} 
\usepackage{amsmath}
\usepackage{amssymb}
\usepackage{booktabs}
\usepackage{multirow}
\usepackage[table]{xcolor}
\usepackage{appendix}

\usepackage[colorlinks]{hyperref}
\AtBeginDocument{\hypersetup{citecolor=teal, linkcolor=teal, urlcolor=teal}}

\usepackage{algorithm}
\usepackage{algpseudocodex}
\algnewcommand\algorithmicinput{\textbf{Input:}}
\algnewcommand\Input{\item[\algorithmicinput]}

\usepackage{subcaption}

\DeclareMathOperator*{\argmaxD}{argmaxD}
\DeclareMathOperator*{\argminD}{argminD}
\DeclareMathOperator*{\mean}{mean}

\newcommand{\card}[1]{\vert#1\vert}

\newcommand{\feat}[1]{\textit{#1}}
\newcommand{\cpt}[1]{{\fontfamily{pcr}\fontsize{8.5pt}{12}\selectfont{\textbf{[#1]}}}}
\newcommand{\cptFeat}[1]{{\fontfamily{pcr}\fontsize{8.5pt}{12}\selectfont{\textbf{#1}}}}

\def\OKFname{\textit{OK}}
\def\KOFname{\textit{KO}}
\def\NWallFname{\textit{NorthWall}}
\def\EWallFname{\textit{EastWall}}
\def\SWallFname{\textit{SouthWall}}
\def\WWallFname{\textit{WestWall}}
\def\ColdFname{\textit{Cold}}
\def\SoundFname{\textit{Sound}}
\def\CaseNbZeroFname{\textit{\#0}}
\def\CaseNbOneFname{\textit{\#1}}
\def\CaseNbTwentyTwoFname{\textit{\#22}}
\def\CaseNbTwentyFourFname{\textit{\#24}}

\def\NDeplFname{\textit{N}}
\def\NEDeplFname{\textit{NE}}
\def\EDeplFname{\textit{E}}
\def\SEDeplFname{\textit{SE}}
\def\SDeplFname{\textit{S}}
\def\SWDeplFname{\textit{SW}}
\def\WDeplFname{\textit{W}}
\def\NWDeplFname{\textit{NW}}
\def\DiagFname{\textit{Diag}}
\def\OrthFname{\textit{Orth}}

\def\FailFname{\textit{Failure}}

\def\L{\textit{\textbf{L}}}
\def\M{\textit{\textbf{M}}}
\def\F{\textit{\textbf{F}}}
\def\I{\textit{\textbf{I}}}
\def\O{\textit{\textbf{O}}}
\def\A{\textit{\textbf{A}}}

\newcommand{\aboveBackwardSTOuF}{\mathit{aboveBackwardSTOuF}}
\newcommand{\aboveSTL}{\mathit{aboveSTL}}
\newcommand{\aboveT}{\mathit{aboveT}}
\newcommand{\activeSynapses}{\mathit{activeSynapses}}
\newcommand{\actResultIneurons}{\mathit{actResultIneurons}}
\newcommand{\addedNeurons}{\mathit{addedNeurons}}

\newcommand{\backwardFiredI}{\mathit{backwardFiredI}}
\newcommand{\backwardFiredO}{\mathit{backwardFiredO}}

\newcommand{\backwardInputSum}{\mathit{backwardInputSum}}
\newcommand{\backwardInputSumL}{\mathit{backwardInputSumL}}
\newcommand{\backwardInputSumOuF}{\mathit{backwardInputSumOuF}}
\newcommand{\backwardSTO}{\mathsf{backwardSTO}}
\newcommand{\backwardSTOo}{\mathsf{backwardSTO[o]}}
\newcommand{\boost}{\mathit{boost}}
\newcommand{\boostedinputSumAiAii}{\mathit{boostedinputSumA1A2}}
\newcommand{\boostedInputSumO}{\mathit{boostedInputSumO}}
\newcommand{\boostParamA}{\mathsf{boostParamA}}
\newcommand{\boostParamO}{\mathsf{boostParamO}}

\newcommand{\coefMult}{\mathit{coefMult}}
\newcommand{\conf}{\mathit{conf}}
\newcommand{\confi}[1]{\mathit{conf_{#1}}}
\newcommand{\cnx}{\mathit{cnx}}
\newcommand{\cnxLO}{\mathit{cnxLO}}
\newcommand{\cnxMAii}{\mathit{cnxMA2}}
\newcommand{\cnxOAi}{\mathit{cnxOA1}}
\newcommand{\cnxOAiii}{\mathit{cnxOA3}}
\newcommand\currentAThreshold{\mathit{currentAThreshold}}

\newcommand{\departLoc}{\mathit{departLoc}}
\newcommand{\decreasedWeight}{\mathit{decreasedWeight}}

\newcommand{\Fail}{\mathit{Fail}}
\newcommand{\FailureNeuron}{\mathit{Failure}}
\newcommand{\firedA}{\mathit{firedA}}
\newcommand{\firedO}{\mathit{firedO}}
\newcommand{\firedL}{\mathit{firedL}}
\newcommand{\firedSet}{\mathit{firedSet}}
\newcommand{\fwrdCnxSum}{\mathit{fwrdCnxSum}}

\newcommand{\inhibitedIneurons}{\mathit{inhibitedIneurons}}
\newcommand{\inhibitedLneurons}{\mathit{inhibitedLneurons}}
\newcommand{\inputBaseSet}{\mathit{inputBaseSet}}
\newcommand{\inputLO}{\mathit{inputLO}}
\newcommand{\inputMAii}{\mathit{inputMA2}}
\newcommand{\inputOAi}{\mathit{inputOA1}}
\newcommand{\inputOAiii}{\mathit{inputOA3}}
\newcommand{\inputSet}{\mathit{inputSet}}
\newcommand{\inputSum}{\mathit{inputSum}}
\newcommand{\inputSumAi}{\mathit{inputSumA1}}
\newcommand{\inputSumAii}{\mathit{inputSumA2}}
\newcommand{\inputSumAiAii}{\mathit{inputSumA1A2}}
\newcommand{\inputSumO}{\mathit{inputSumO}}

\newcommand{\lastSpikedA}{\mathit{lastSpikedA}}
\newcommand{\lastSpikedO}{\mathit{lastSpikedO}}
\newcommand{\lastSpikesVector}{\mathit{lastSpikesVector}}
\newcommand{\learnAT}{\mathsf{learnAT}}
\newcommand{\learningA}{\mathit{learningA}}
\newcommand{\learningO}{\mathit{learningO}}
\newcommand{\learningSet}{\mathit{learningSet}}
\newcommand{\LF}{\mathit{LF}}
\newcommand{\LR}{\mathit{LR}}

\newcommand{\maxCoefModulatedCnx}{\mathsf{maxCoefModCnx}}
\newcommand{\maxLastSpike}{\mathit{maxLastSpike}}
\newcommand{\maxLearn}{\mathit{maxLearn}}
\newcommand{\minCoefModulatedCnx}{\mathsf{minCoefModCnx}}
\newcommand{\modulatedCnx}{\mathit{modulatedCnx}}
\newcommand{\modulatedCnxOAi}{\mathit{modulatedCnxOA1}}
\newcommand{\modulatedCnxMAii}{\mathit{modulatedCnxOA2}}
\newcommand{\MF}{\mathit{MF}}
\newcommand{\missedFeatures}{\mathit{missedFeatures}}
\newcommand{\move}{\mathit{move}}
\newcommand{\MotAct}{\mathit{MotAct}}

\newcommand{\newLoc}{\mathit{newLoc}}
\newcommand{\newLocFeatures}{\mathit{newLocFeatures}}
\newcommand{\newSynapses}{\mathit{newSynapses}}
\newcommand{\noiseA}{\mathsf{noiseA}}
\newcommand{\noiseO}{\mathsf{noiseO}}

\newcommand{\predFeatIneurons}{\mathit{predFeatIneurons}}
\newcommand{\predictedFeatures}{\mathit{predictedFeatures}}
\newcommand{\predictions}{\mathit{predictions}}
\newcommand{\predictionErrors}{\mathit{predictionErrors}}
\newcommand{\prevFiredO}{\mathit{prevFiredO}}
\newcommand{\probaNewSynapses}{\mathit{probaNewSynapses}}
\newcommand{\probaNewSynapsesLO}{\mathit{probaNewSynapsesLO}}
\newcommand{\probaNewSynapsesOAi}{\mathit{probaNewSynapsesOA1}}
\newcommand{\probaNewSynapsesMAii}{\mathit{probaNewSynapsesMA2}}
\newcommand{\probaNewSynapsesOAiii}{\mathit{probaNewSynapsesOA3}}

\newcommand{\randomNeuron}{\mathit{randomNeuron}}
\newcommand{\remainsToDecrease}{\mathit{remainsToDecrease}}
\newcommand{\RemoveByKey}{\mathit{RemoveByKey}}

\newcommand{\selectedMode}{\mathit{selectedMode}}
\newcommand{\SGR}{\mathit{SGR}}
\newcommand{\shortTermMemoryCoef}{\mathit{shortTermMemoryCoef}}
\newcommand{\spikingO}{\mathit{spikingO}}
\newcommand{\spikingL}{\mathit{spikingL}}
\newcommand{\STFail}{\mathsf{STFail}}
\newcommand{\STL}{\mathsf{STL}}
\newcommand{\STO}{\mathsf{STO}}
\newcommand{\STOo}{\mathsf{STO[o]}}
\newcommand{\suitable}{\mathit{suitable}}
\newcommand{\synapsesToDelete}{\mathit{synapsesToDelete}}

\newcommand{\TnbFiredA}{\mathsf{TnbFiredA}}
\newcommand{\TnbFiredO}{\mathsf{TnbFiredO}}
\newcommand{\TnbLearningA}{\mathsf{TnbLearningA}}
\newcommand{\TnbLearningO}{\mathsf{TnbLearningO}}
\newcommand{\TnbQueryA}{\mathsf{TnbQueryA}}
\newcommand{\TnbQueryO}{\mathsf{TnbQueryO}}

\newcommand{\undecided}{\mathit{undecided}}
\newcommand{\unsuitable}{\mathit{unsuitable}}

\newcommand{\WSA}{\mathsf{WS_{A}}}
\newcommand{\WSO}{\mathsf{WS_{O}}}

\def\backwardSpikeO{\textsc{backwardSpikeO}}
\def\backwardSpikeIQuery{\textsc{backwardSpikeI\_Query}}
\def\backwardSpikeL{\textsc{backwardSpikeL}}
\def\boostInputLO{\textsc{boostInputLO}}

\def\calculateInputSumsA{\textsc{calculateInputSumsA}}
\def\calculateProbaNewSynapsesA{\textsc{calculateProbaNewSynapses\_A}}
\def\calculateProbaNewSynapsesLO{\textsc{calculateProbaNewSynapses\_LO}}

\def\calculateNewLoc{\textsc{calculateNewLoc}}
\def\calculateLR{\textsc{calculateLR}}
\def\chooseAMove{\textsc{chooseAMove}}
\def\computeBackwardInputSumsOuF{\textsc{computeBackwardInputSumsOuF}}

\def\learnA{\textsc{learnA}}
\def\learnO{\textsc{learnO}}

\def\makeAChoice{\textsc{makeAChoice}}
\def\makeAStep{\textsc{makeAStep}}
\def\makePredictions{\textsc{makePredictions}}
\def\modulateCnx{\textsc{modulateCnx}}

\def\query{\textsc{query}}

\def\randomChoice{\textsc{randomChoice}}
\def\ratePredictions{\textsc{ratePredictions}}
\def\replaceAllSynapses{\textsc{replace\_All\_Synapses}}
\def\replaceAllInactiveSynapses{\textsc{replace\_AllInactive\_Synapses}}
\def\replInactPlusSomeActSynapses{\textsc{replace\_AllInactive+SomeActive\_Synapses}}
\def\replaceSomeInactiveSynapses{\textsc{replace\_SomeInactive\_Synapses}}

\def\selectLearningNeuronsA{\textsc{selectLearningNeurons\_A}}
\def\spikeA{\textsc{spikeA}}
\def\spikeO{\textsc{spikeO}}
\def\sumInputsAiAii{\textsc{sumInputsA1A2}}

\def\updateLastSpikes{\textsc{updateLastSpikes}}

\title{\textsf{\textbf{SNN-based online learning of concepts and action laws in an open world}}
\author{\textsf{\textbf{Christel Grimaud}}\,\thanks{Corresponding author: christel.grimaud@irit.fr} , \textsf{\textbf{Dominique Longin, Andreas Herzig}}\\ \normalsize{\textsf{Université de Toulouse, CNRS, Toulouse INP, UT3, IRIT, France}}}
\date{}
}

\begin{document}
%

%

\maketitle

\begin{abstract}
\textsf{We present the architecture of a fully autonomous, bio-inspired cognitive agent built around a spiking neural network (SNN) implementing the agent's semantic memory. This agent explores its universe and learns concepts of objects/situations and of its own actions in a one-shot manner. While object/situation concepts are unary, action concepts are triples made up of an initial situation, a motor activity, and an outcome. They embody the agent's knowledge of its universe's action laws. Both kinds of concepts have different degrees of generality. To make decisions the agent queries its semantic memory for the expected outcomes of envisaged actions and chooses the action to take on the basis of these predictions. Our experiments show that the agent handles new situations by appealing to previously learned general concepts and rapidly modifies its concepts to adapt to environment changes. }
\end{abstract}


\section{Introduction}

The ability of a cognitive agent to act adequately in a given environment depends on its ability to predict how performing a given action will affect its current situation; 
that is, it depends on the agent's knowledge of the laws of the universe that determine the effect of its possible actions on its environment (hereafter the \emph{action laws} of the universe). 
How artificial agents can acquire these laws and how these should be updated if their environment changes has proved a difficult question. In the case where the intended environment is \emph{open}---that is, where the agent's designer cannot foresee all the situations the agent might encounter in the future---, providing a suitable set of action laws to the agent  ``by hand'' is unfeasible. The only viable solution is that the agent continuously learns the relevant laws from experience, just as natural agents (humans and animals) do. 
Crucially, this learning process should allow for generalization over disparate experiences, so that the agent is able to behave appropriately in new situations. It should also allow for rapid modification of the learned action laws to accommodate environment changes. Finally, it should address the fact that however big its memory resources may be they will always be finite, while the number of distinct objects and situations it may encounter in an open world is unbounded.

\smallskip
The aim of the present paper is to show how such a learning (autonomous, online, 
achieving generalization and rapid updating while being adapted to open universes), could be done. Solving this problem would be useful in applications such as mobile autonomous robots  (e.g. service robots) living in open worlds.

The proposed approach's main hypothesis is that natural agents' ability to perform well in our open and changing world largely relies on the fact that they store their knowledge in the form of rapidly updatable \emph{concepts} with various degrees of generality. 
Presumably, natural agents first form concepts about the encountered objects/situations, and then use these as elements for composing more complex concepts, notably concepts of action. The latter constitutes their knowledge of their environment's action laws.
Concepts are a very compact way to store information, as more general concepts can apply to a whole range of distinct cases while being usable in new similar situations. They also allow for efficient updating, as the adjustment of a single concept modifies all the inferences and further decisions that can be made on its basis.
Another important hypothesis is that natural agents' endless ability to learn new concepts despite having finite brains relies on some efficacious management of forgetting.Forgetting is inevitable for an agent with finite memory living in an open world; but it needs to avoid  \emph{catastrophic forgetting}, that is, some kind of forgetting that would severely affect its ability to act judiciously in its current world.
Most likely, natural agents avoid catastrophic forgetting by selectively forgetting their less important knowledge (such as, e.g., details and old, unused memories) 
and reallocating the corresponding neurons to encode new useful knowledge.   
We suggest that artificial agents could take inspiration from these strategies and use some artificial neural network to learn and store concepts, and query this network to make predictions about the outcome of envisaged actions.

\smallskip
To test this idea, we here build an artificial agent with an SNN at its core. This agent lives in a very simple virtual world, composed of rooms which may be, or not, accessible (hence, knowable) to it. At first, the agent is confined to one single room and learns by itself how to act in it according to its own interests. Then at some point a door opens to a new room containing some never encountered before objects and situations. Yet, although these are new to the agent, some general laws are preserved from one room to the other. Our experiments show that having learned these laws in the first room allows the agent to act by and large properly in the second one, as soon as it enters it. They also show that the agent is able to learn new laws holding in the second room without catastrophic loss of previous knowledge. Finally, some changes are introduced in the second room, rendering some of the previously learned rules obsolete while some new rules become true. Again, experiments show that the agent quickly updates its knowledge to account for these changes.

\smallskip
The paper is organized as follows. Section \ref{sec: Related works} discusses related work and Section \ref{sec: The agent and its universe}  presents the agent and its universe. Section \ref{sec: Implementing ...} gives a general overview of the neural network,  while Section \ref{sec: NN functioning} details its functioning. Section  \ref{sec:  agent's functioning} describes the agent's workings and Section  \ref{sec: Results} presents the experiments conducted on the agent and discusses the results. Section  \ref{sec: Conclusion} concludes and outlines future possible developments of the framework.

\section{Related works}\label{sec: Related works}

The research problem addressed in this work is autonomous online learning, updating and generalization of concepts and action laws in an open universe. 
To our knowledge, no existing approach addresses this problem in all its dimensions, but these are investigated in separate research fields.

\smallskip
\textbf{Continual Learning} tackles the problem of lifelong knowledge acquisition \cite{WanEtAl24,Les20}. Its main challenge is to avoid catastrophic loss of previous knowledge when acquiring new knowledge; a secondary research axis is one-shot/few-shots learning, which is the ability to learn online from one or few examples \cite{WanEtAl20}. 
However, current approaches mostly consider the learning of \emph{tasks} (mostly image classification/recognition tasks, but also some more complex tasks such as playing games \cite{KirEtAl17}), not of concepts nor action laws. 
Furthermore, most of them rely on learning algorithms that use labelled training data (mainly some supervised learning algorithms), which makes them unsuitable for open world autonomous agents.

\smallskip
\textbf{Concept Learning} has mainly been studied in view of explainability \cite{Gup2024}, mostly of classification models (e.g.,  \cite{Koh2020}) but also of decision making in the context of reinforcement learning \cite{DasEtAl2023,ZabEtAl2023}. 
For this reason, many proposals are dedicated to learning a human-predefined set of concepts using some annotated data. In the field of Image Classification some approaches deal with the extraction of concepts from data \cite{WanEtAl2022,Gho19,Has2019}, but then these are extracted from labelled classes of images, which again is unsuitable for open world autonomous agents.
The vast majority of these methods also disregards the hierarchical organization of concepts from particular to more general, and they generally do not address one-shot learning, online revision or updating of concepts.

\smallskip
\textbf{Action Laws Learning} has been studied from various perspectives. In Dynamic Epistemic Logic,  \cite{Bol18} proposed a method to learn an action model through successive observations of transitions between states. But this method does not address generalization nor environment changes and only considers \emph{universally applicable} actions (i.e., actions that can be executed in every logically possible state), a condition real-world actions rarely satisfy. 
In the field of Planning, \cite{BonEtAl2019} showed how to learn abstract actions from a few carefully chosen instances of some general planning problem. But said instances come with their own set of ground actions which must be known beforehand, so this approach cannot be used in open worlds, where an agent needs to incrementally learn from experience. 

\smallskip

\textbf{Reinforcement Learning} (RL) is concerned with agents learning by experience how to achieve a goal in an optimal manner. A number of variants have been developed, which all have in common that the agent learns through trials and errors by means of a reward system, and that what is finally learned is a \emph{policy}, i.e., a function which takes a given state of the agent as an argument and returns an action (or a set of actions) to be taken.  
RL approaches have proved very successful in a wide range of domains, and for this reason they are quite popular.
However, they struggle to adapt to environment changes and to revise the learned policies \cite{Kir2023,Far2018}. 
This could be related with the fact that in natural agents, learning policies is rather a matter of \emph{procedural memory} (``memory of procedures''), a memory system that typically requires a large number of learning trials \cite{Tul85, KnoEtAl17}. 
By contrast, the \emph{semantic memory}, which stores facts about the world in the form of concepts, is a fast learning system that can learn or update a concept over one single experience.
It thus seems that an artificial agent that would be able to learn and update in real time a conceptual model of its environment and to use it to make its decisions would be able to quickly adapt to environment changes. 

RL approaches also typically encounter difficulties in contexts where rewards are scarce or deceptive \cite{Har2019,Oca2023}, which is the case of most real world contexts. Although some solutions have been proposed (see notably  \cite{Eco2021}), dealing with sparse rewards is still an active field of research in RL. 
But since concept learning does not depend on rewards, an agent able to learn concepts and to use them to make its decisions would be immune to the problem.

\smallskip
\textbf{In the field of robotics}, semantic information has long been recognized as instrumental in autonomous robots' decision making \cite{Pra2020}, but there have been very few attempts to make  robots capable of learning semantic knowledge from experience.
The most prominent are probably \cite{WanEtAl2016} and  \cite{Nas2019}, which both use fusion ART  (for \emph{Adaptive Resonance Theory)} neural networks to encode concepts.
However these neural networks rely on the extension of their architecture to learn new knowledge (i.e., new neurons are created to encode new concepts), which is unsuitable for autonomous agents with finite memory resources living in an open world. 

\bigskip
Agents living in complex open environments (such as the real world) can be confronted with a huge amount of information in the course of their life. Retaining the totality of that information and further processing it to use it does not seem a practical option, as this would be extremely costly in terms of memory, computational power and energy consumption. 
This is especially true in the case of agents such as autonomous mobile robots, for which frugality regarding these same resources can be critical.
 For these agents, memorizing the whole of the experienced objects and events cannot be taken as a reasonable goal or criterion. Rather, forgetting appears as the unavoidable counterpart of continual learning, and the question is how to manage it so as to preserve the agent's performance as much as possible. 
The most natural way to do this is by ensuring that the previous knowledge that is lost through new learning is the agent's less useful knowledge at the time of the learning. How to characterize ``less useful knowledge'' and how to preferentially select it for deletion is a key issue to address to efficiently manage forgetting.
In turn, such an efficient management of forgetting appears as a necessary condition for making lifelong learning autonomous agents in complex open environments possible.

\bigskip
The present paper aims at providing a proof of concept of a cognitive agent satisfying the above requirements. As a first step, it focuses on the learning of concepts and only uses them in some basic decision making to demonstrate the agent's learning abilities. 
However the agent is designed so as to allow the retrieval of the learned concepts and action laws, which could then be encoded in symbolic format and used in more complex decision making involving explicit reasoning and action planning. The implementation of such complex reasoning abilities in the agent is left to future work.

\bigskip
In the absence of studies investigating the considered research problem in all its dimensions, comparing the proposed setup with existing approaches seems inappropriate. Indeed, this would only take into account some of the dimensions of the problem and fail to evaluate the proposal relative to its goal. 
It should also be remarked that the standard metrics commonly used to assess neural networks' learning performance cannot be used in the present case.
Notably, notions such as  \emph{Average Accuracy}, \emph{Forward} and \emph{Backward Transfer} \cite{Lop2017, Cha2018} used in the field of Continual Learning to assess learning accuracy, generalization and resistance to catastrophic forgetting cannot be used. 
Indeed, these metrics were devised to assess the performance of classifiers and are built upon a basic notion of \emph{accuracy} that consists in checking whether two objects (generally the predicted class and the actual class of a sample item) are identical. 
But in the present proposal a prediction is not a single object but a set of objects (namely a set of predicted features), which is to be compared with another set of objects (the actual situation's set of features). Simply checking whether these two sets are identical would be a rather crude way to assess the prediction's accuracy: a more fine-grained comparison is needed, based on set inclusion properties. 
For these reasons, a suitable set of experiments and metrics was devised (see Section \ref{sec: Results}).
%

\section{The agent and its universe}\label{sec: The agent and its universe}

This section describes the universe the agent lives in and gives a general overview of the agent. It also clarifies the notions of concepts and action laws used in the sequel.


\subsection{The universe}

\smallskip
The universe is designed so as to support a set of experiments intended to assess the agent’s abilities. It presents a number of challenges for the agent to face, while being kept as simple as possible to allow the precise monitoring of the agent's performances. 

\begin{figure}[t]
 \centering
 \includegraphics[width=1\columnwidth]{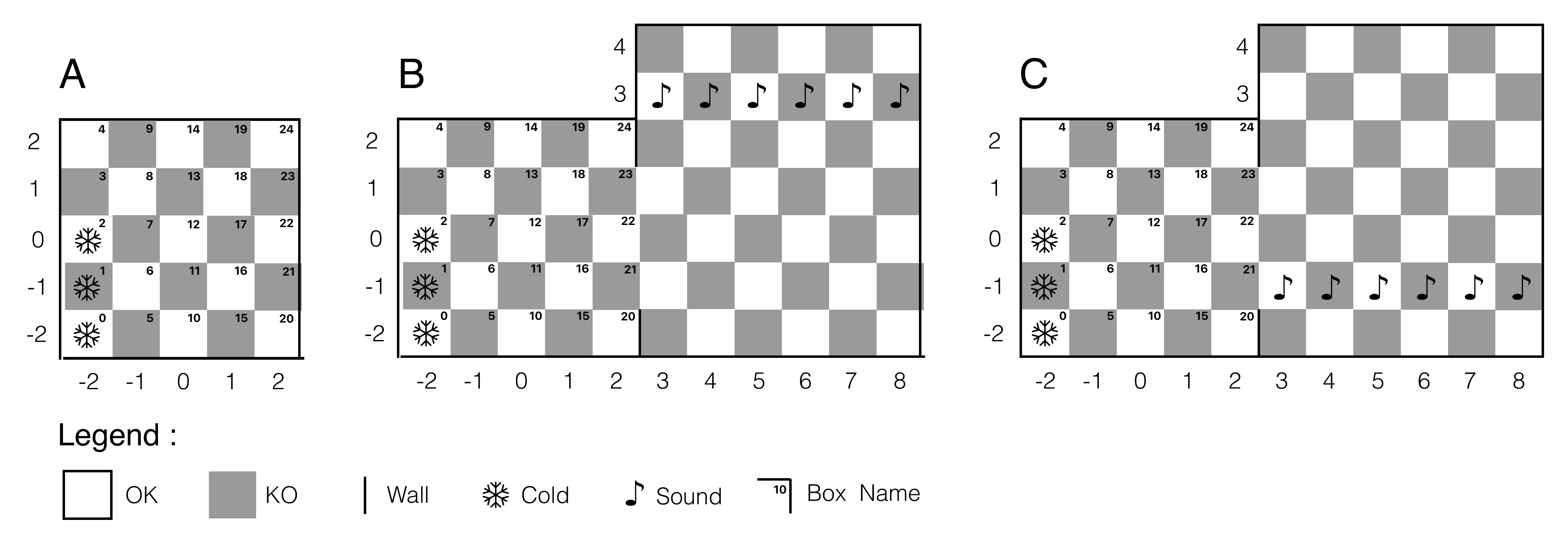}
 \caption{The agent's accessible world. A: in the first phase, Room 1 only; B: after opening the door, Rooms 1 and 2. C : after the sound feature is moved downwards (from row 3 to row -1) in Room 2.}
 \label{fig: Universe}
 \end{figure}

\smallskip
The universe is built over a grid of boxes, which we (not the agent) identify using an orthonormal coordinate system (see Figure~\ref{fig: Universe}). Each box represents a particular location in the agent's universe and possesses a particular set of features the agent is able to perceive, drawn from the set  $\LF\ = \{$\OKFname, \KOFname, \NWallFname, \EWallFname, \SWallFname, \WWallFname, \ColdFname, \SoundFname,  \CaseNbZeroFname, \CaseNbOneFname, ..., \CaseNbTwentyFourFname $ \}$. 
For example, the box with coordinates $(-2, -2)$ has the feature set  $\LF_{(-2, -2)} = \{$\OKFname, \SWallFname, \WWallFname, \ColdFname, \CaseNbZeroFname $ \}$, while the box with coordinates $(5, 0)$ has the feature set  $\LF_{(5, 0)} = \{$\KOFname$\}$. For $n \in \{1, ..., 24\}$, the expression ``\textit{\#n}'' is to be taken as a particular name for a box. 
It is therefore a feature, and not all boxes need to have one. Two boxes with the same feature set are indistinguishable for the agent.
Some features are mutually exclusive, which means that they cannot pertain at the same time to a same box. Mutually exclusive features are \OKFname\ and \KOFname\ on the one hand, and boxes' names on the other. This means that a box cannot have both \OKFname\ and \KOFname\ in its feature set, and can have at most one name.

The rooms are made out of boxes, and delimited with impassable walls. For instance, Figure~\ref{fig: Universe}.A shows the first room, which is composed of 25 boxes. Opening a door amounts to removing the wall features from the corresponding boxes' feature sets (as in Figure~\ref{fig: Universe}.B, where the  \EWallFname\ feature has been removed from boxes $(2, -1)$, (2, 0) and  $(2, 1)$).
More generally, changes in boxes' feature sets model environment changes. For example, at the time the door opens the boxes  $(3, 3)$ to  $(8, 3)$ all have \SoundFname\ in their respective feature sets, but at some point the \SoundFname\ feature moves to the boxes  $(3, -1)$ to  $(8, -1)$ (Figure~\ref{fig: Universe}.C).

When considering the agent's universe, at any point in time we only consider the boxes to which it has access, the set of which is always finite. Note that this does not contradict the unboundedness of the universe: although each possible room is finite (as are rooms in real world), the number of possible rooms the agent may discover in its life and the number of new objects it may encounter in them, as well as the number of changes that may occur in these rooms, are unbounded.
Being able to handle these novelties (react appropriately and update its knowledge on the fly) is the agent's main challenge.
\smallskip

As regards the features, \KOFname\ corresponds to some unpleasant stimulus the agent spontaneously wants to avoid, and \OKFname\ to the absence of such a stimulus. The other features convey some indifferent information. 
It should be stressed that \OKFname\ and \KOFname\  are not rewards in the sense of RL, as they play no role in the neural network's learning process (see section \ref{sec: NN functioning} for details). Their only purpose is to motivate the agent's choices and to allow us to assess its ability to make appropriate decisions. 

\smallskip
The distribution of OK and KO boxes, which is common to both rooms, provides some general (non-monotonic) laws of the universe (such as \emph{``going North-East from an OK box leads to another OK box''}), the learning of which is the agent’s second challenge.
Boxes' names in the first room are specific features (i.e., they apply to one single object). They are used to check that the agent also forms particular concepts of individual boxes and uses them to learn particular rules (such as \emph{``going North-East from the OK box with name `\#12' leads to the OK box with name `\#18'}). 
\ColdFname\  is a feature common to a small number of boxes. It is mainly used to increase the number of features boxes may have (from two to five), and also to check that the agent can form concepts with an intermediate degree of generality such as, e.g., the concept of cold OK boxes with a West Wall (see Section \ref{sec: Concepts and action laws} for a formal definition of generality).
Being able to handle concepts with various degrees of generality is a third challenge.
\SoundFname\  is used to test the agent's ability to use the learned general rules in the presence of novelty, and also to check that it can learn new concepts and action laws without suffering catastrophic forgetting. Notably, after the door opens (Figure~\ref{fig: Universe}.B) the agent is expected to learn new general rules such as \emph{``going North from a box with sound leads to a box with a north wall''.} Moving the \SoundFname\ feature from its ``up'' position (Figure~\ref{fig: Universe}.B) to its ``down'' position (Figure~\ref{fig: Universe}.C) is used to test the agent's ability to update its knowledge when the environment changes.

\subsection{The agent}
\begin{figure}[t]
 \centering
 \includegraphics[width=.8\columnwidth]{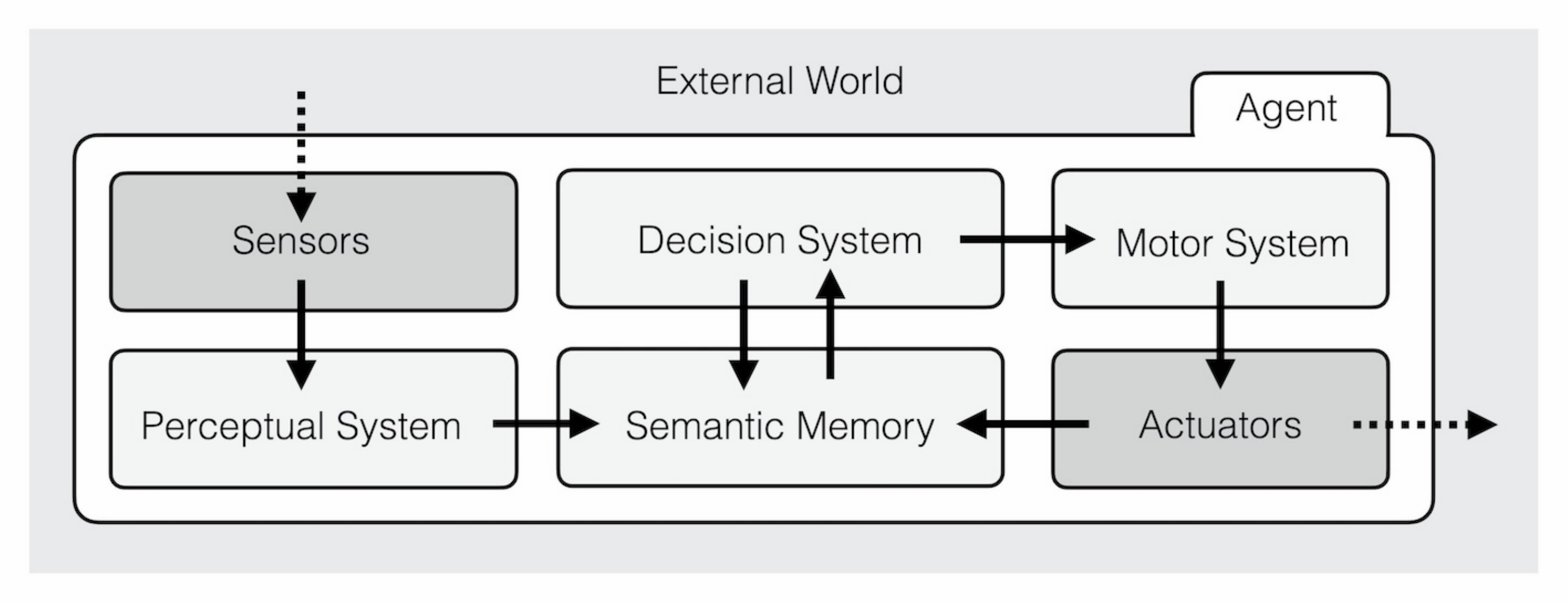}
 \caption{Schema of the agent}
 \label{fig:Schema Agent}
 \end{figure}

The agent is composed of a set of sensors, a perceptual system, a semantic memory, a decision system, a motor system and a set of actuators (see Figure~\ref{fig:Schema Agent}). 
Sensors collect data from the external world and feed it to the perceptual system, which performs feature/object recognition. Neural networks doing this while relying on unsupervised learning from unlabelled data already exist (e.g., \cite{ThiEtAl18,Khe17}), so we simply suppose that the agent's perceptual system operates as intended and provides the semantic memory with the appropriate inputs, namely, the correct and complete set of features of the agent's current location. 
Semantic memory forms concepts by binding together sets of features, and stores them for further retrieval. Its modeling is the main focus of the paper. 
The decision system is the other important part: it queries the semantic memory to predict the outcome of possible actions, and decides which one to take on the basis of these predictions. This decision is then sent to the motor system, which activates the actuators to perform the corresponding motor activity. Information from the actuators is sent back to semantic memory through proprioception, allowing the agent to memorize the motor-related features of the realized actions.\smallskip

The agent's possible actions consist in steps from one box to another adjacent box, in any of the eight directions. Formally, an action is a triple made up of a depart location, a motor activity, and an outcome. By ``motor activity'' we mean the fact that the agent's actuators are activated so as to make it move to the immediate next box in the selected direction. The set of the agent's possible motor activities is therefore the set $\MotAct = \{$\textit{North, North-East, East, South-East, South, South-West, West, North-West}$\}$.
The set of motor activity features the agent is able to perceive by proprioception is the set $\MF\ = \{$\NDeplFname, \NEDeplFname, \EDeplFname, \SEDeplFname, \SDeplFname, \SWDeplFname, \WDeplFname, \NWDeplFname, \DiagFname, \OrthFname$\}$, where the first eight are specific to each particular motor activity, while \DiagFname\ and \OrthFname\ are more general features shared by all motor activities yielding diagonal/orthogonal moves. 
In cases where there is a wall at the edge of the depart box in the selected direction, the agent bumps into it and remains at the same place. We then say that the action's outcome is a failure. Otherwise, the action's outcome is the agent's new location. 
%

%
\subsection{Concepts and action laws}\label{sec: Concepts and action laws}
The agent can form two kinds of concepts. First, concepts of ``things'', in the broad sense. These concepts bind together co-occurrent features and can be seen as some sort of conjunction in which conjuncts have different ``weights'', reflecting the fact that some features are more important than others in a concept's definition \cite{Fre08}.
They store the agent's knowledge about locations and more generally about any object, so we call them \emph{object concepts}. The second kind is relational concepts. These take other concepts as elements and bind them together into tuples. Concepts of actions are of this kind: they bind together the agent's concepts of a depart location, a performed motor activity, and a subsequent outcome, in the order in which they were experienced.  

\smallskip
An object concept X  is said to be \emph{general}, as opposed to \emph{particular}, if there is another concept Y such that the set of features composing X is a strict subset of the set of features composing Y. Y is then said to be \emph{more particular} than X. These definitions capture the fact that if X is more general than Y then the set of objects X applies to is a superset of the set of objects Y applies to.
In line with this idea, in this work an action concept is said to be \emph{general} if the object concept of its initial situation is general or its motor activity component only contains \DiagFname\ or \OrthFname\ (thus, if X and Y are action concepts such that X is more general than Y, then the set of actions X applies to is a superset of the set of actions Y applies to).
Generality of both kinds of concepts is understood relatively to the set of concepts the agent possesses at some point, so no concept is general or particular in itself. 

For example, when visiting the box $(0, 0)$ the agent may form the particular object concept \cpt{OK,\#12}, which is a memory of an OK place with name \#12, and only applies to this particular box in its accessible universe. 
If the agent then moves North-East and arrives at box $(1, 1)$, it can form the particular object concept \cpt{OK,\#18}, and also the particular action concept \cpt{\cpt{OK,\#12},\cpt{NE,Diag},\cpt{OK,\#18}} which corresponds to the memory of being in an OK place with name \#12 and then moving North-East to arrive at another OK place with name \#18.
Yet, after visiting a number of locations having the feature \feat{OK} in common, the agent may also form the general object concept \cpt{OK}. 
Furthermore, it is a general rule in its accessible universe that moving North-East from an OK location always leads to another OK location, except for when there is a wall at the North or East edge of the depart box. Therefore, after having experienced a number of North-East moves from various OK locations, the agent may form general action concepts such as \cpt{\cpt{OK},\cpt{NE,Diag},\cpt{OK}} and  \cpt{\cpt{OK,NorthWall}\cpt{NE,Diag},\cpt{Failure}}.
Such general concepts capture the general (non-monotonic) action laws of the agent's universe. 
It is on them that the agent shall rely on to behave in never encountered situations. 

It should be mentioned that this representation of action laws is slightly different from the one used in the AI field of planning (see e.g.\ \cite{DBLP:journals/ai/FikesN71}), which involves a notion of \emph{executability} that specifies the conditions that need to hold for an action to be possible. 
Yet the notion of executability can readily be recovered from action concepts by stating that the action represented by the action concept \cpt{x,\,y,\,z} is executable (up to the agent's knowledge) in a situation $s$ if $s$ satisfies all the features composing \cptFeat{x} and \mbox{\cptFeat{z} $\neq$ \cpt{Failure}.}

\section{Implementing the agent's semantic memory in the neural network: overview}\label{sec: Implementing ...}

This section presents the neural network implementing the agent's semantic memory. It first gives an overview of SNNs, and more particularly of the neural models and the learning rules which are relevant for that purpose. Then it presents the network's general architecture.

\subsection{Spiking Neural Networks}\label{sec: Spiking Neural Networks}
Spiking Neural Networks (SNNs) are artificial neural networks in which information is transmitted between neurons by the means of spikes fired by neurons at other neurons. At each time step the activation level of each neuron is computed as a function of both its previous activation level and the amount of activation received by the neuron at this time step. If this activation level reaches a certain  threshold (the neuron's \emph{spike threshold}) the neuron fires a spike and its activation level is reset to some base value. Emitted spikes are conveyed from one neuron to another through weighted connections that modulate the amount of input transmitted by spikes to the receiving neuron. Learning consists in adjusting these weights. 

SNNs are well suited for autonomous learning in open worlds because they allow for \emph{Spike Time Dependent Plasticity} (STDP), a family of biologically plausible learning rules which can achieve unsupervised online learning from unlabelled data \cite{ThiEtAl18}. There are a number of STDP variants, but all have in common that the modification of the connection's weight between two neurons depends on the relative timing of their spikes. In the most popular version, the connection between two neurons is reinforced if the input neuron spikes just before the output neuron; it is decreased if it spikes just after the output neuron; and it is left unchanged otherwise. 

There are also many different implementations of STDP  in SNNs, ranging from highly detailed and biologically accurate models to drastically simplified ones. Simplified versions are known for being rapid and energy-efficient \cite{YamEtAl2022}, which are interesting properties for autonomous robots. 

The JAST learning rule \cite{ThoEtAl19,Tor23} is certainly one of the simplest STDP implementations one can find. In this family of models, connections between neurons are binary (i.e., weights are either 0---meaning no connection---or 1) and each neuron has a fixed number of incoming connections. Learning is achieved through the replacement, at each time step, of a number $n_{swap}$ of inactive connections (i.e., connections from currently inactive input neurons) with an equal number of connections from unconnected active input neurons.
Each neuron $i$ has two different thresholds $T_{fire}$ and $T_{learn}(i)$, which respectively decide whether the neuron spikes and learns.  $T_{fire}$ is fixed and is the same for all neurons, while $T_{learn}(i)$ is initialized at  some minimal value and increased each time $i$ learns until it reaches $T_{fire}$. It is then fixed at this same value, so $T_{learn}(i) \leq T_{fire}$ in any case. 
The number $n_{swap}(i)$ of connections to replace is initialized at some maximal value and is then decreased each time $i$ learns, until $n_{swap}(i) = 0$ when $T_{learn}(i) = T_{fire}$. 
This mechanism ``freezes'' the neuron after learning, thus  preventing it from forgetting the acquired knowledge. 
In another version of JAST the spiking threshold  $T_{fire}$ is not fixed but is dynamically adjusted so that a number $n$ of neurons (the \emph{``n-best''} neurons) spike at each step.

Although this learning rule is appealing for its simplicity, it cannot be used as it is for the present purpose. 
Obviously, freezing neurons is not suitable for continual learning and updating. 
Furthermore, neurons having both binary connections and a fixed number thereof would be unable to encode concepts having a variable number of features, which is required since an object's number of features may vary over time 
(for instance, when the door is closed the box $(2, 0)$ has the feature set  $\LF_{(2, 0)} = \{$\OKFname,  \CaseNbTwentyTwoFname, \EWallFname$\}$, but this last feature is lost when the door opens). 
More generally, online learning and updating in a changing world presents challenges to neural networks that seem difficult to overcome while sticking to JAST's simple setup. 

For this reason, the neural network implementing the agent’s semantic memory only retains some of the JAST learning rule's ideas, while relying on a more refined modeling of the neurons' internal dynamics to address these issues.
Retained ideas are mainly the use of binary synapses, and the fact that learning consists in swapping a number of these so that the sum of the connections' weights on any given neuron remains constant through learning. This character ensures that there can be no ``dead'' neurons, i.e., neurons having lost most of their incoming connections and which are therefore unable to respond to any input. 
However here a same input neuron may have multiple synapses onto a same output neuron, which means that the connections' weights are in fact  integers. 
Another retained idea is the use, in some specific cases, of dynamic spiking thresholds, to ensure that a given input always has a number of neurons responding to it. 

The paper's original contribution regarding SNNs mainly concerns the use of some internal metrics of the neurons to regulate learning and modulate information transmission. For instance, the input neurons' ability to establish new synapses onto output neurons depends on the number of synapses they already have, which facilitates the learning of rare or new features (such as e.g. \SoundFname). 
Another example is that the choice of the neurons recruited for learning a new concept depends on the time of their last spike. This allows to preferentially select less used neurons for new learning and to prevent in this manner the forgetting of more used---and therefore presumably more useful---knowledge. 
A third example is the modulation of information transmission between neurons in the querying process, which depends on the weight sum of input neurons' connections onto output neurons. This modulation favours the transmission of more specific information over that of less specific information, which is critical in non-monotonic contexts (see Section \ref{sec:Querying}). 
It should be remarked that these methods do not depend on the specific type of spiking neurons used in the proposal and could thus be applied to regulate learning and forgetting and to balance information retrieval 
in other contexts.

%
%
%
\subsection{The neural network's architecture}\label{Architecture}
\begin{figure}
 \centering
 \includegraphics[width=\textwidth]{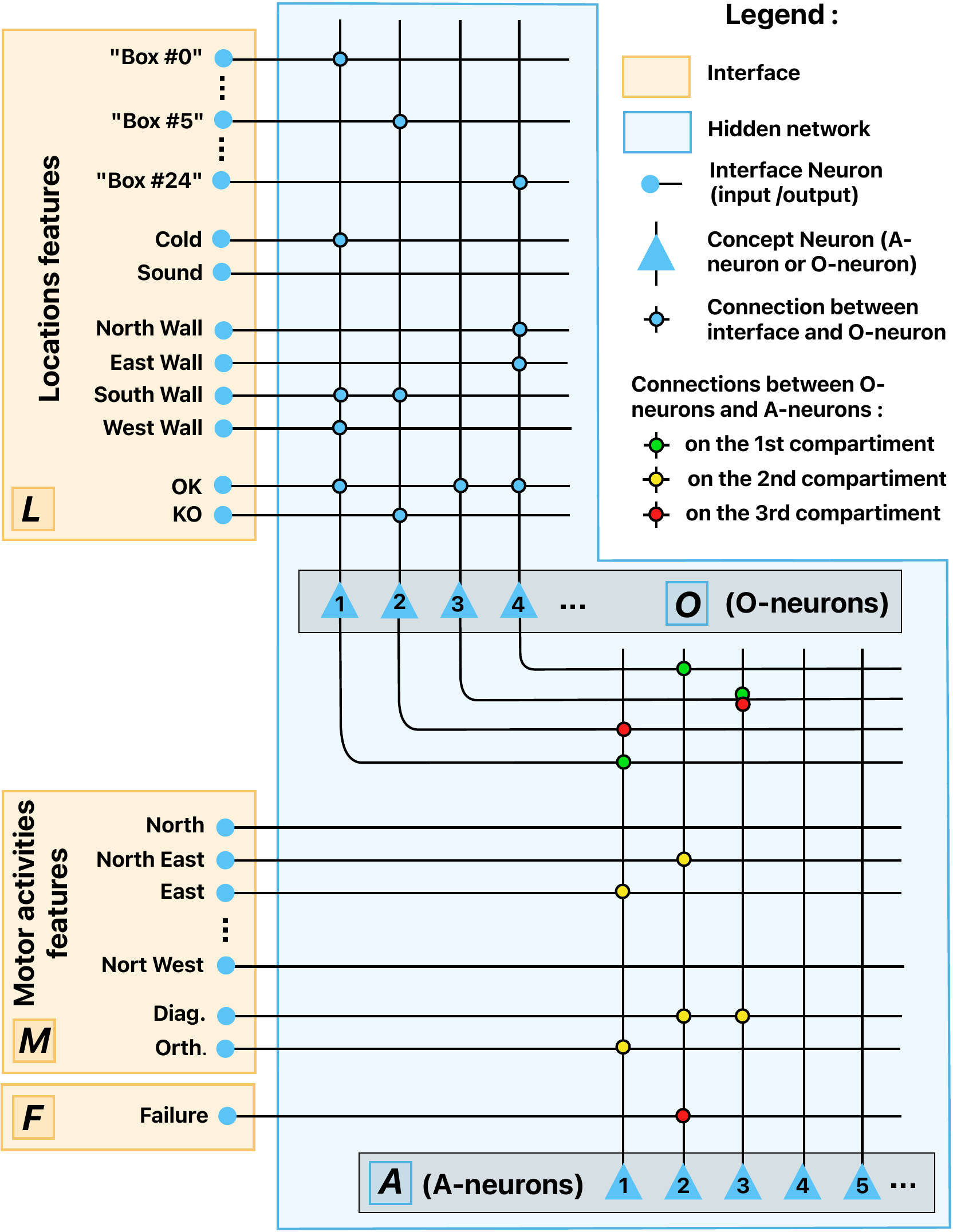}
 \caption{Schema of the SNN. O-neurons \#1 and  \#2 respectively encode the concepts \cpt{\#0,\,Cold,\,SouthWall,\,WestWall, \,OK} and  \cpt{\#5,\,SouthWall,\,KO}. A-neuron \#1 encodes the concept  \cpt{\cpt{\#0,\,Cold,\,SouthWall,\,WestWall,\,OK},\,\cpt{E,\,Orth}, \,\cpt{\#5,\,SouthWall,\,KO}}. A-neuron \#3 encodes the concept  \cpt{\cpt{OK},\,\cpt{Diag},\,\cpt{OK}}.  For graphical readability, connections weights are omitted. }
 \label{fig:SNN}
 \medskip
 \end{figure}

The network is composed of an interface, which communicates with the agent's other components, and a body of hidden neurons which is itself divided into two layers (see Figure~\ref{fig:SNN}). 
The first layer learns object concepts and the second learns action concepts. For this reason we call their neurons, respectively, \emph{object concept neurons} (O-neurons for short) and \emph{action concept neurons} (A-neurons). 
 This architecture draws on neuroanatomical studies according to which concepts are represented in the brain by hierarchically organized \emph{concept neurons}, each receiving information from some lower neurons and sending reciprocal connections to these same neurons so that it can reactivate them for information retrieval \cite{Qui12,BauEtAl21,Shi10}. 
 For simplicity these reciprocal connections are not modelled as such, but instead information is allowed to flow in both directions along the same connections: from interface neurons to O-neurons and from there to A-neurons for learning and querying, and the other way round for retrieving information. A key point is that interface neurons are both input and output neurons, depending on the phase of the computation. 

 Interface neurons (I-neurons for short) mainly support the representation of features, be it of the visited locations or of the agent's own motor activities. An additional neuron acts as a failure detector, specifically firing when the agent bumps into a wall and remains at the same place. All of them have their labels fixed from the start. 
I-neurons encoding mutually exclusive features have fixed reciprocal inhibitory connections which prevents them from firing together in a same computing process: as soon as one has fired, the others are shut down for the rest of the process. The $\FailureNeuron$ neuron has reciprocal inhibitory connections with each and any of the L-neurons.

The first layer of hidden neurons (O-neurons) is composed of 100 neurons with a differentiated dynamics depending on whether their input source is I-neurons or A-neurons. O-neurons learn co-occurrences of features.

The second layer is composed of 400 compartment neurons with three separate input compartments. The first compartment receives connections from O-neurons, the second from motor activities I-neurons, and the third from O-neurons and the $\FailureNeuron$ neuron. 
Inputs received at each compartment are unable to trigger a spike  by themselves, but the first and second compartments respectively make their next compartment ready to receive and transmit inputs for a certain amount of time.  In this manner, inputs can only be efficient if they occur in the correct order, so that A-neurons encode sequences of inputs. The use of compartment neurons to learn sequences of inputs was suggested in \cite{Cui16,HawEtAl16}.
%

\section{The neural network's functioning}\label{sec: NN functioning}
%
This section describes the neural network's functioning and details its implementation. For readability, inessential algorithms and parameters settings are relegated in Appendices.

\subsection{General notions}

Let \L,  \M, and \F\ be the sets of interface neurons  that respectively encode locations' features, motor activities' features, and failure (note that \F\ is the singleton $\{\FailureNeuron \}$). \mbox{\I\ = \L\ $\cup$ \M\ $\cup$ \F} is the set of interface neurons. 
The sets of O- and  A-neurons are respectively noted \O\ and  \A . For convenience the neurons from \L\ are called \emph{``L-neurons''}, and similarly for the other sets.

\subsubsection{Connections.}
Synapses between any two neurons always have a weight of $1$, but an input neuron may have multiple synapses onto a same output neuron, so the weight of a connection between two neurons is a positive integer. By a slight abuse of language, when there is no synapse connecting two neurons we say that the weight of their connection is 0.
Connections between the above sets of neurons are stored in separate arrays $\cnxLO$, $\cnxOAi$, $\cnxMAii$ and $\cnxOAiii$, where $\cnxLO$ reads \textit{``connections from \L\ to \O''},  $\cnxOAi$ reads \textit{``connections from \O\ to \A's first compartment''}, and similarly for the others. $\cnxLO[l][o]$ denotes the weight of the connection between the L-neuron $l$ and the O-neuron $o$, that is, the number of synapses that compose this connection.

These arrays are initialized at random, in such a way that for any O-neuron $o$ 
\begin{align*}
\sum \cnxLO[l][o]~\textrm{for}~l \in \L &= \WSO,
 \intertext{and for any  A-neuron $a$}
\sum \cnxOAi[o][a]~\textrm{for}~o\in \O  &= \WSA,\\
\sum \cnxMAii[m][a]~\textrm{for}~m \in \M &= \WSA,\\
\sum \cnxOAiii[n][a]~\textrm{for}~n \in \O \cup \F&=  \WSA,
\end{align*}
where $\WSO = 40$ and $\WSA = 30$. These values were chosen so as to allow enough redundancy between synapses to ensure robustness of learning while keeping the computational complexity as low as possible.
The agent's learning consists in the update of these arrays.

\subsubsection{Spike thresholds.}
Each O-neuron $o$ has two spike thresholds. The first one is noted $\STOo$ and governs its forward spiking, that is, the cases where its received input originates from L-neurons. The second is noted $\backwardSTOo$ and governs  its backward spiking, that is, the cases where the input originates from A-neurons. Both spike thresholds are fixed. $\STOo$ is a random integer taken in $\{22, ..., 31\}$, and $\backwardSTOo = \mathrm{floor}(\STOo/3)$. These spike thresholds are stored in separate vectors $\STO$ and $\backwardSTO$ of length $\card{\O}$.
Differentiated forward spike thresholds favour the encoding of concepts with different degrees of generality, as neurons with lower spike thresholds tend to spike more easily in response to inputs that do not perfectly fit their connections, which brings them to learn more general concepts. Lower backward spike thresholds eases information retrieval. 
L-neurons all have the same spike threshold $\STL = 50$, and the $\FailureNeuron$ neuron has its own spike threshold $\STFail = 40$.
 Spike thresholds for A-neurons are dynamically adjusted, in a way that is described in Section \ref{sec: A-neurons} below.

\subsubsection{Last spikes.}
The number of steps performed by the agent since the last spike of each O- and A-neuron is stored in separate vectors  $\lastSpikedO$ and $\lastSpikedA$ of respective lengths $\card{\O}$ and $\card{\A}$. These vectors are initialized at random with values comprised between
1 and some upper bound. 
Last spike values are then increased by 1 at each step and reset to 0 each time the corresponding neuron spikes. 
Only the spikes that occur after a learning episode induce a reset, as they are the ones that indicate a successful encoding of the input. Notably, spikes occurring in the course of the querying process do not induce a reset.

\subsubsection{Synapses' Growth Rates.}
While the number of incoming synapses on O- and A-neurons is strictly constrained by $\WSO$ and $\WSA$, the number of outgoing synapses from I- and O- neurons is softly constrained by \emph{Synpases Growth Rates} ($\SGR$s). 
For any neuron $n$ from \I\ or \O, $n$'s $\SGR$ is a  decreasing function of the number of outgoing synapses $n$ has at the moment. 
 $\SGR$s account for the fact that neurons cannot indefinitely grow new synapses, hence that a neuron having already a lot of synapses should be less prone to establish new ones than a neuron having only a few of them (see Algorithms \ref{calculateProbaNewSynapsesLO} and \ref{calculateProbaNewSynapsesA} below for details).
 $\SGR$s are instrumental in avoiding the over-representation of frequent inputs over rare ones, and so in ensuring that the agent is able to learn new or rarely observed features, objects or actions.

\subsubsection{Implementation apparatus.}
In the algorithms provided in the sequel we introduce and use dictionary structures in which each value $v$ can be accessed through an index (\emph{``key''}) $k$. Formally, a dictionary can be seen as a set of pairs $(k, v)$ such that for any pairs $(k, v)$ and $(k', v')$ in $D$, $k \neq k'$. 
If $D$ is a dictionary, then the operation $D[k] \leftarrow  v$ means that the value $v$ is added at index $k$ in $D$.  If $k$ already exists in $D$  then its value is overwritten. 
If $S$ is a subset of $D$'s indices, the operation $\mathit{RemoveByKey}(D, S)$ modifies $D$ by removing all its elements having their indices in $S$. That is, $\mathit{RemoveByKey}(D, S) = D \setminus \{ (k,v) \in D  \mid k \in S \}$. We also use a function $\argmaxD$ which given a dictionary $D$ returns the subset of its indices with maximal values (that is, if $x = \max(\{v  \mid  (k,v) \in D\})$, then $\argmaxD(D) = \{k  \mid  (k,x) \in D\}$).

We also use Python's Numpy function \randomChoice$(a, n, \mathit{replace}$, $p)$\label{randomChoice}, where $a$ is a set, a vector or an array, and the other parameters are optional with default values:
 \begin{compactitem}
    \item[-] $n$ is the number of elements to return (default is $1)$.
    \item[-]  $\mathit{replace}$ specifies whether the chosen elements in $a$ should be replaced between successive draws when $n > 1$  (default is $\mathrm{True}$).
    \item[-] $p$ is a probability distribution over $a$ (default is the the uniform distribution).
\end{compactitem} 
 \randomChoice\ returns a vector of length $n$ of randomly chosen elements from $a$, following the probability distribution $p$. If $n$ is not specified, \randomChoice\ does not return a vector $[v]$ but a single value $v$.

\subsection{O-neurons}\label{sec: O-neurons}
%

\subsubsection{Forward firing of O-neurons.}
The set of L-neurons that fire in response to an input from either the perceptual system (if the agent is observing its current location) or the decision system (if the agent is querying its semantic memory) is noted $\inputLO$.  The set of O-neurons that fire in response to $\inputLO$\ is noted $\firedO$. It is computed as follows (see Algorithm \ref{spikeO}).
\begin{algorithm}[t]
\caption{Function \spikeO\ (forward spiking of O-neurons)}\label{spikeO}
\begin{algorithmic}[1]
\Input $(\inputLO, \cnxLO)$
\LComment{Note that: \\
    $\inputLO \subseteq \L$,\\
    $cnxLO$ is a $(\card{\L} \times \card{\O})$ matrix of integers, \\
    $\STO$ is a vector of integers of size $\card{\O}$,\\
    $\TnbFiredO$ is a constant (the maximum desired number of firing neurons).}
\State $\firedO \gets \emptyset$ \Comment{the set of neurons that fired}
\State $\inputSumO \gets [0, ..., 0]$ \Comment{vector of integers of size $\card{\O}$}\\
    \Comment{we note $\inputSumO[o]$ the value of $\inputSumO$ at position $o$}
\State $\aboveT \gets \emptyset$ \Comment{dictionary with keys in \O\ and values in $\mathbb{N}$}      
\For{$o \in$ \O}
    \State $\inputSumO[o] \gets \sum_{l \in \inputLO}\cnxLO[l][o]$ 
        \Comment{amount of activation received by neuron o}
    \If{$\inputSumO[o] > \STOo$}
          \State $\aboveT[o] \gets  \inputSumO[o] - \STOo $
    \EndIf
\EndFor
\While{$\card{\firedO} < \TnbFiredO \textbf{ and } \aboveT \neq \emptyset$}
    \State   $S \gets \argmaxD(\aboveT)$ \Comment{the set of O-neurons with maximal over threshold activation value}
    \State $\firedO \gets \firedO \cup S$ 
    \State $\aboveT \gets \RemoveByKey(\aboveT, S)$
\EndWhile
\Statex \Return  $(\inputSumO,\, \firedO)$
\end{algorithmic}
\end{algorithm}
First, for any neuron $o$ in \O\ the sum $\inputSumO[o]$ of inputs received by $o$ is computed. 
O-neurons for which $\inputSumO[o]$ is above their spike threshold $\STOo$ then fire in turn, starting by those with highest difference with their spike threshold, until a target number $\TnbFiredO$  of firing neurons is reached or no neuron reaching its spike threshold remains. $\TnbFiredO$ is the maximum desired number of firing O-neurons; it serves to keep the number of firing O-neurons under control.
 
\subsubsection{Information retrieving.}
Given a set $\firedO$ of firing O-neurons, the information stored by these neurons can be retrieved by the backward firing of L-neurons. This consists in having the O-neurons from $\firedO$ send an input to L-neurons to make them spike  (see Algorithm \ref{backwardSpikeL} in Appendices). The process is similar as above, except that in order to suppress noise only the connections having a weight above a noise threshold $\noiseO$ transmit the input. Furthermore, mutual inhibition between \L-neurons encoding mutually exclusive features makes that once a \L-neuron encoding a feature $f$ has fired, all the \L--neurons encoding features $f'$ excluded by $f$ are prevented from firing.

\subsubsection{Learning of O-neurons.}
Learning of object concepts  (see Algorithm \ref{learnO}) occurs after each observation by the agent of its current situation. The set of O-neurons selected for learning is noted $\learningO$. These are primarily the neurons that fired at observation time, that is, the neurons from $\firedO$. 
%
\begin{algorithm}[t]
\caption{Function \learnO\ (O-neurons' learning)}\label{learnO}
\begin{algorithmic}[1]
\Input $(\inputLO, \inputSumO, \firedO, \cnxLO, \lastSpikedO)$
\LComment{Recall that: \\ 
    $\inputLO \subseteq \L$ and $firedO \subset \O$; \\
    $\inputSumO$ and $lastSpikedO$ are vectors of integers of size $\card{\O}$; \\
    $\cnxLO$ is a $(\card{\L} \times \card{\O})$ matrix of integers;\\
    $\TnbLearningO$ is a constant (the minimum desired number of learning O-neurons).}
\State $\learningO\ \gets \firedO$
\If {$\card{\learningO} < \TnbLearningO$}
    \State $\boostedInputSumO \gets \hyperref[boostInputLO]{\boostInputLO} (\learningO, \inputSumO, \lastSpikedO)$
    \While{$\card{\learningO} < \TnbLearningO \textbf{ and } \boostedInputSumO \neq\emptyset$} 
        \State $\addedNeurons \gets \argmaxD(\boostedInputSumO)$ 
            \Comment{subset of $\O$}
        \State $\learningO \gets \learningO \cup \addedNeurons$
        \State $\boostedInputSumO \gets \RemoveByKey(\boostedInputSumO, \addedNeurons)$ 
       \EndWhile
\EndIf
\State $\probaNewSynapsesLO \gets$ \hyperref[calculateProbaNewSynapsesLO]{\calculateProbaNewSynapsesLO}$(\inputLO,$ $ \cnxLO)$

 \State $\firedL \gets \hyperref[backwardSpikeL]{\backwardSpikeL}(\firedO, \cnxLO)$ \Comment{$\firedL$ is a set of firing L-neurons}

\If{$\inputLO \subseteq \firedL$}
        \State $\cnxLO \gets \hyperref[replaceAllInactiveSynapses]{\replaceAllInactiveSynapses} (\learningO, \inputLO, \cnxLO, \L, $ $\probaNewSynapsesLO)$
\Else
    \If{$\card{\learningO} < \TnbFiredO$}
        \State $\boostedInputSumO \gets \hyperref[boostInputLO]{\boostInputLO} (\learningO, \inputSumO, \lastSpikedO)$ 
        \State $\learningO \gets \learningO \cup \argmaxD(\boostedInputSumO)$   
    \EndIf
    \State $\maxLastSpike \gets \max(\{\lastSpikedO[o] \mid o \in \learningO\}\})$
    \State $\maxLearn \gets \hyperref[randomChoice]{\randomChoice}(\{o \in \learningO \mid \lastSpikedO[o] = \maxLastSpike)$ 
    \State{$S \gets \learningO\setminus \{\maxLearn\}$} \Comment{subset of $\learningO$}
    \State $\cnxLO \gets$ \hyperref[replaceAllInactiveSynapses]{\replaceAllInactiveSynapses} $(S, \inputLO, \cnxLO, \L, $ $\probaNewSynapsesLO)$ \Comment{same as in the previous case for neurons in $S$}
    \State $\cnxLO \gets$ \hyperref[replaceAllSynapses]{\replaceAllSynapses} $(\maxLearn, \inputLO, \cnxLO, \L, $ $\probaNewSynapsesLO, \WSO)$ \Comment{replace all synapses, active or inactive, for $\maxLearn$}
\EndIf
\Statex \Return $\cnxLO$
\end{algorithmic}
\end{algorithm}
However, it may happen that there are not enough of them, which is materialized by the fact that a target number $\TnbLearningO$ of learning neurons is not reached. This typically occurs when the observed situation is new to the agent, as in such a case very few neurons may be able to reach their spike threshold hence to fire. 

When this happens, the network seeks additional neurons to make learn by ``boosting'' the input received by O-neurons (see Algorithm \ref{boostInputLO}). 
\begin{algorithm}[t]
\caption{Function \boostInputLO\ (boosts the input from \L\ to O-neurons)} \label{boostInputLO}
\begin{algorithmic}[1]

\Input $(\learningO, \inputSumO, \lastSpikedO)$
\LComment{Recall that: \\
    $\learningO \subseteq \O$; \\
    $\inputSumO$ and $\lastSpikedO $ are vectors of integers of size $\card{\O}$.}
\State $\boostedInputSumO \gets \emptyset$  \Comment{dictionary with keys in \O\ and values in $\mathbb{R}$}

\For{$o \in \O \setminus \learningO$}
    \State $\boost \gets  (\lastSpikedO[o] + \boostParamO) / \boostParamO )$
    \Comment{$\boostParamO = 50$ is a constant; it determines the slope of the function's curve.}
    \State $\boostedInputSumO[o] \gets \inputSumO[o] * \boost$
\EndFor
\Statex \Return $\boostedInputSumO$

\end{algorithmic}
\end{algorithm}
In practice,  for each non-firing O-neuron a boosting coefficient $\boost$ is calculated, which is an increasing function of $\lastSpikedO[o]$. $o$'s boosted input sum $\boostedInputSumO[o]$ is then obtained by multiplying $\inputSumO[o]$ by this coefficient. 
Neurons with the highest boosted input sums are then successively added  to $\learningO$ until $\TnbLearningO$ is reached or no more neurons can be found. As neurons that did not spike for a long time receive more boosting, they tend to be preferentially recruited in this process. This means that the network preferentially selects neurons encoding less used knowledge to learn new information.
Of course, these neurons will lose part of their previously stored information in this new learning process, bringing the agent to forget some of its previous knowledge. But this will only affect its least used knowledge, as these neurons were the least used at the time of the learning.
To ensure that boosting only occurs when very few O-neurons fire in response to the input, $\TnbLearningO$  is set to $\TnbFiredO/2$ ($\TnbFiredO =12$ and $\TnbLearningO = 6$).
The learning of O-neurons essentially consists in replacing synapses they have with currently non-firing L-neurons with synapses from currently firing L-neurons (i.e., those from $\inputLO$). 
To regulate the formation of new synapses and avoid the over-representation of frequently observed features, a probability distribution $\probaNewSynapsesLO$ is defined based on L-neurons' \emph{Synapses Growth Rates}  (see Algorithm \ref{calculateProbaNewSynapsesLO}). 

\begin{algorithm}[t]
\caption{\calculateProbaNewSynapsesLO\ (calculates L-neurons' probability to establish new synapses onto O-neurons)}\label{calculateProbaNewSynapsesLO}
\begin{algorithmic}[1]
\Input $(\inputLO, \cnxLO)$
\LComment{Recall that: \\
    $\inputLO \subseteq \L$;  \\
    $\cnxLO$ is a $(\card{\L} \times \card{\O)}$ matrix of integers.}
\State $\SGR, \probaNewSynapsesLO \gets \emptyset$ 
    \Comment{dictionaries with keys in $\inputLO$ and values in $\mathbb{R}$.}
\For{$l \in \inputLO$}
    \State $\fwrdCnxSum \gets \sum_{o \in \O} \cnxLO[l][o]$
    \State $\SGR[l] \gets \tanh( (-\fwrdCnxSum  + a) / b ) + c )$
    \LComment{$a = 300$, $b = 150$ and $c = 2$ are constants;  they determine the horizontal offset,  the slope  and the vertical offset of the function's curve.}
\EndFor
\State $s \gets \sum_{l \in \inputLO} \SGR[l]$
\For{$l \in \inputLO$}
    \State $\probaNewSynapsesLO[l] \gets \SGR[l] / s$
\EndFor

\Statex \Return $\probaNewSynapsesLO$

\end{algorithmic}
\end{algorithm}

The learning process then depends on the accuracy of the agent's knowledge relative to the current situation.
To assess it, the O-neurons from $\firedO$ send a backward input to L-neurons, and the resulting set $\firedL$ of firing L-neurons  (see Algorithm \ref{backwardSpikeL} in Appendices) is compared with the initial input $\inputLO$. This corresponds to the agent comparing what it observes (represented by $\inputLO$) with what it would have expected based on its current knowledge (represented by $\firedL$). 
If $\inputLO \subseteq \firedL$, that is, if all the observed features can be retrieved from the firing O-neurons, then
for any $o \in \learningO$, all of $o$'s inactive synapses are deleted and replaced with new synapses from neurons from $\inputLO$, following the probability distribution $\probaNewSynapsesLO$ (see Algorithm \ref{replaceAllInactiveSynapses} in Appendices).
This procedure tends to reinforce O-neurons' connections with L-neurons encoding well-shared features at the expense of connections with L-neurons encoding more specific features, and is the driving force for the formation of general object concepts.

If, on the contrary,  $\inputLO \not \subseteq \firedL$, then the network first checks if $\card{\learningO}$ reaches the target number $\TnbFiredO$ (recall that $\TnbFiredO > \TnbLearningO$). If not,  it looks for additional neurons for $\learningO$, using the same boosting method as above except that only neurons with the maximal boosted input sum are added (most of the time, this amounts to one single neuron). 
The learning process is then similar as before, except that for one O-neuron with maximal $\lastSpikedO$ value all synapses are replaced  whether active or inactive (see Algorithm \ref{replaceAllSynapses} in Appendices). 
This particular neuron will therefore have the opportunity to establish connections with all the neurons from $\inputLO$ and will thus tend to learn the particular situation with all its features, counterbalancing the general trend towards generalization. The choice of a neuron with a maximal $\lastSpikedO$ value for this complete re-learning is meant to mitigate the forgetting of useful knowledge.
%

\subsection{A-neurons}\label{sec: A-neurons}

\subsubsection{ Inputs.}
The sets of neurons from \O, \M\ and \O\ $\cup$ \F\ that send inputs to A-neurons' first, second and third compartments are respectively noted $\inputOAi$, $\inputMAii$ and $\inputOAiii$.  
Inputs on the first compartment encode an initial situation, inputs on the second compartment encode motor activity features, and inputs on the third compartment encode an action's outcome.
Such inputs may occur in two different situations: when the neural network learns about a just performed action, or when the decision system queries the neural network for the expected outcome of an envisaged action (see Section \ref{sec:Querying}). 

%
\subsubsection{A-neurons' forward spiking.}
Let $\inputOAi$ be a set of firing O-neurons and $\inputMAii$ a set of firing M-neurons. The set of A-neurons that fire in response to the sequence of inputs $(\inputOAi,\allowbreak \inputMAii)$ is noted $\firedA$. It is computed as follows  (see Algorithm \ref{spikeA}). 
\begin{algorithm}[t]
\caption{\spikeA\ (forward spiking of A-neurons)}\label{spikeA}
\begin{algorithmic}[1]
\Input $(\inputOAi, \inputMAii, \cnxOAi, \cnxMAii)$
\LComment{Note that: \\
    $\inputOAi \subset \O$ and $\inputMAii \subset \M$; \\
    $\cnxOAi$ and $\cnxMAii$ are matrices of integers of respective dimensions $(\card{\O}\times\card{\A})$ and \mbox{$(\card{\M}\times\card{\A})$;} \\
    $\TnbFiredA$ is a constant (the desired number of firing A-Neurons).}
\State $\inputSumAi \gets$  \hyperref[calculateInputSumsA]{\calculateInputSumsA}$(\inputOAi, \cnxOAi )$
\State $\inputSumAii \gets$  \hyperref[calculateInputSumsA]{\calculateInputSumsA}$(\inputMAii, \cnxMAii )$
\State $\inputSumAiAii \gets$ \hyperref[sumInputsAiAii]{\sumInputsAiAii}$(\inputSumAi, \inputSumAii)$
\State $V \gets \{v \mid  (a, v) \in \inputSumAiAii \}$
\State $\firedA \gets \emptyset$\Comment{set of A-neurons that fired}
\While{$\card{\firedA} < \TnbFiredA$ and $V \neq \emptyset$}
    \State $\currentAThreshold \gets  \max(V)$
    \State $\firedA \gets \{a \in \A \mid \inputSumAiAii[a]\geq  \currentAThreshold \}$
    \State $V \gets V \setminus \{\currentAThreshold\}$ 
\EndWhile
\Statex \Return $(\inputSumAiAii,\, \firedA)$
\end{algorithmic}
\end{algorithm}
%
%
%
First, the sums of inputs received by each A-neuron $a$ on its first two compartments are separately computed (see Algorithm \ref{calculateInputSumsA} in Appendices). Then, if both compartments receive significant input (i.e., above some noise threshold $\noiseA$) these inputs add up and are transmitted to the neuron, otherwise they are discarded (see Algorithm \ref{sumInputsAiAii} in Appendices).
The set $\firedA$ is obtained by first setting A-neurons' spike threshold to the highest input sum value (so only A-neurons with the highest input sum fire), and then repeatedly lowering it to the next highest value until the number of firing A-neurons reaches a target number $\TnbFiredA$. 
Each time the spike threshold is lowered all the A-neurons with input sums above the threshold spike again, so A-neurons with higher inputs sums spike repeatedly. This behaviour will be used in the querying process described in Section \ref{sec:Querying} below. 
%

\subsubsection{A-Neurons' learning.}
Learning of action concepts (see Algorithm \ref{learnA}) takes place after each step made by the agent. 
In this case, $\inputOAi$  is the set $\firedO$ resulting from the agent observing its depart location and $\inputMAii$ is the set of M-neurons firing by proprioception as the agent performs its motor activity.
\begin{algorithm}[t!]
\caption{\learnA\ (A-neurons' learning)}\label{learnA}
\begin{algorithmic}[1]
\Input$(\inputOAi, \cnxOAi, \inputMAii, \cnxMAii, \inputOAiii, \cnxOAiii,$ $ \actResultIneurons, \predFeatIneurons, \lastSpikedA)$
\LComment{Recall that: \\
    $\inputOAi \subset \O$, $\inputMAii \subset \M$, and $\inputOAiii \subset \O$ or $\inputOAiii = \F$; \\
    $\cnxOAi$, $\cnxMAii$ and $\cnxOAiii$ are matrices of integers of respective dimensions $(\card{\O} \times \card{\A})$, $(\card{\M} \times \card{\A})$ and $(\card{\O \cup \F} \times \card{\A})$; \\
    $\actResultIneurons \subset \L$ or $\actResultIneurons = \F$; \\
    $\predFeatIneurons\subset \L$ or $\actResultIneurons = \F$; \\
    $\lastSpikedA$ is a vector of integers of length $\card{\A}$.}
\State $\learningA \gets$ \hyperref[selectLearningNeuronsA]{\selectLearningNeuronsA}$(\inputOAi, \inputMAii, \cnxOAi, \cnxMAii, \lastSpikedA)$
\State \mbox{$\probaNewSynapsesOAi \gets$ \hyperref[calculateProbaNewSynapsesA]{\calculateProbaNewSynapsesA}$(\inputOAi, \cnxOAi)$}
\State \mbox{$\probaNewSynapsesMAii \gets$ \hyperref[calculateProbaNewSynapsesA]{\calculateProbaNewSynapsesA}$(\inputMAii, \cnxMAii)$}
\State \mbox{$\probaNewSynapsesOAiii \gets$ \hyperref[calculateProbaNewSynapsesA]{\calculateProbaNewSynapsesA}$(\inputOAiii, \cnxOAiii)$}
\State $\LR \gets$  \hyperref[calculateLR]{\calculateLR}$(\learningA, \lastSpikedA)$
\State $\predictionErrors \gets\predFeatIneurons\setminus\actResultIneurons$
\State $\missedFeatures \gets \actResultIneurons \setminus \predFeatIneurons$

\If{$\predictionErrors = \missedFeatures  = \emptyset$}
    \State $\cnxOAi \gets$ \hyperref[replaceAllInactiveSynapses]{\replaceAllInactiveSynapses} $(\learningA, \inputOAi,$ $ \cnxOAi, \O, \probaNewSynapsesOAi)$
    \State $\cnxMAii \gets$ \hyperref[replaceAllInactiveSynapses]{\replaceAllInactiveSynapses} $(\learningA, \inputMAii, $ $\cnxMAii, \M, \probaNewSynapsesMAii)$
    \State $\cnxOAiii \gets$ \hyperref[replaceSomeInactiveSynapses]{\replaceSomeInactiveSynapses} $(\learningA, \inputOAiii,$ $ \cnxOAiii, \probaNewSynapsesOAiii, \LR)$ 

\Else
    \If{$\predictionErrors \neq \emptyset$}
        \State $\cnxOAi \gets$  \hyperref[replInactPlusSomeActSynapses]{\replInactPlusSomeActSynapses} $(\learningA, \inputOAi, \cnxOAi, \O, \probaNewSynapsesOAi, \LR)$
        \State $\cnxMAii \gets$  \hyperref[replInactPlusSomeActSynapses]{\replInactPlusSomeActSynapses} $(\learningA, \inputMAii, \cnxMAii, \M, \probaNewSynapsesMAii, \LR)$
        \State$ \cnxOAiii \gets$ \hyperref[replaceSomeInactiveSynapses]{\replaceSomeInactiveSynapses} $(\learningA, $ $ \inputOAiii, \cnxOAiii, \probaNewSynapsesOAiii, \LR)$ 

    \EndIf
    \If{$\missedFeatures \neq \emptyset$}
        \State$ \cnxOAi \gets$ \hyperref[replInactPlusSomeActSynapses]{\replInactPlusSomeActSynapses} $(\learningA, \inputOAi, \cnxOAi, \O, \probaNewSynapsesOAi, \LR)$
        \State $\cnxMAii \gets$ \hyperref[replInactPlusSomeActSynapses]{\replInactPlusSomeActSynapses} $(\learningA, \inputMAii, \cnxMAii, \M, \probaNewSynapsesMAii, \LR)$
        \State $\cnxOAiii \gets$ \hyperref[replInactPlusSomeActSynapses]{\replInactPlusSomeActSynapses} $(\learningA, \inputOAiii, \cnxOAiii, \O \cup \F, \probaNewSynapsesOAiii, \LR)$
    \EndIf
\EndIf

\Statex \Return $ (\cnxOAi, \cnxMAii, \cnxOAiii)$  

\end{algorithmic}
\end{algorithm}
$\actResultIneurons$ is the set of interface neurons that encode the action's outcome. If the action failed (that is, if the agent bumped into a wall) $\actResultIneurons$ is the set \F\ $= \{\FailureNeuron \}$; otherwise it is the set $\inputLO$ of L-neurons encoding the new location's features. 
$\inputOAiii$ is either  the set $\firedO$ of O-neurons firing in response to $\inputLO$, or  \F\ if the action failed.
$\predFeatIneurons$ is the set of interface neurons that encode the agent's prediction about the action's outcome. It results from  the agent querying its memory for the action's expected outcome before taking it (see Algorithm \ref{makeAStep} in Section \ref{sec:makeAStep}). If the agent predicted a failure then $\predFeatIneurons = \F$, otherwise  $\predFeatIneurons\subset \L$.

The set of A-neurons selected for learning is noted $\learningA$. To obtain it (see Algorithm \ref{selectLearningNeuronsA} in Appendices), the network first gets a number of  A-neurons to fire in response to the input sequence $(\inputOAi, \inputMAii)$.
However, A-neuron's dynamic spiking threshold makes it possible that A-neurons with only very weak connections from these input neurons ---but with stronger connections from other input neurons--- fire. Recruiting them for learning could thus lead to the loss of valuable previous knowledge. To avoid this, only A-neurons whose input sum reaches a learning threshold $\learnAT$  are selected for learning. 
If the number of such neurons reaches a target number $\TnbLearningA$, the network directly proceeds to learn. Otherwise, additional neurons are selected by boosting the input sums received by A-neurons in a way similar to the one used for O-neurons.  

For each input set $i$ from $\{\inputOAi, \inputMAii, \inputOAiii\}$, a probability distribution \linebreak$\probaNewSynapses_i$ is then computed  (see Algorithm \ref{calculateProbaNewSynapsesA} in Appendices). It  models the propensity of each neuron $n \in i$ to grow new synapses to A-neurons' corresponding compartment. This probability is based on input neurons' $\SGR$s, just as in O-neurons' case.

For each neuron $a$ in $\learningA$, a learning rate $\LR(a)$ is computed (see Algorithm \ref{calculateLR} in Appendices). It represents $a$'s propensity to modify its connections through learning, by specifying how many of its synapses should be at most/at least replaced in the learning process.  
This number  is calculated in such a way that A-neurons with the highest $\lastSpikedA$ value get the maximal learning rate $\WSA$, those with the lowest $\lastSpikedA$ value get a learning rate of $1$, and the others get learning rates in between. 
The idea behind this is that neurons having not spiked for a long time should be more ready to modify their connections so to adjust to the precise current situation and thus get to encode particular concepts, while neurons that often spike should minimally change their connections each time they spike so to tune themselves to the common features shared by all these situations and thus get to encode general concepts.

The learning process then depends on the accuracy of the agent's expectations relative to the action's outcome. 
If these expectations were both correct (i.e., all predicted features are actually present) and complete (i.e., all actually present features were predicted), then for any learning neuron $a$ the learning process deletes all of $a$'s inactive synapses in the first two compartments, but only some of them in the third one  (see Algorithms \ref{replaceAllInactiveSynapses} and \ref{replaceSomeInactiveSynapses} in Appendices). 
More precisely, the number of inactive synapses to delete is $\min(\LR(a), n)$, where $n$ is the number of inactive synapses on $a$'s third compartment. 
Then, in all three compartments new synapses replacing the deleted ones are chosen at random according to the probability distribution $\probaNewSynapses_i$.

If the agent's expectations were not correct, then for any learning neuron $a$ not only all of $a$'s inactive synapses are deleted and replaced on the first two compartments, but also a number of randomly chosen active synapses (see Algorithm \ref{replInactPlusSomeActSynapses} in Appendices). Here the total number of synapses to be replaced is $\max(\LR(a), n)$. The learning process for $a$'s  third compartment is the same as in the first case.
As previously for O-neurons, replacing some active synapses in addition to the inactive ones allows these compartments to ``re-specialize'' to the input. This helps the neuron to only fire in response to this precise action, hence to retain more accurate information about its outcome.

If the agent's expectations were not complete, then for any learning neuron $a$ the number of replaced synapses is $\max(\LR(a), n)$ in all three compartments. If the predictions were neither correct nor complete, the learning process successively runs the two above procedures.

\section{The agent's functioning}\label{sec:  agent's functioning}

This section first explains how the agent queries its semantic memory to predict the outcomes of envisaged actions, and then how the agent uses these predictions to make its decisions. Finally, it describes  the whole process of making a step.

\subsection{Querying the network}\label{sec:Querying}

To query the neural network, the decision system first sends an input to the L-neurons that encode the features of the considered initial situation, bringing a set $\inputLO$ of L-neurons to fire. This triggers the firing of a set $\firedO$ of O-neurons, following the process described in Section \ref{sec: O-neurons}. $\firedO$ acts as an input for A-neurons' first compartment, that is, $\inputOAi = \firedO$. 
Given an envisaged motor activity $\move$, the decision system then sends an input to the M-neurons that encode $\move$'s features. These fire, sending an input  $\inputMAii$ to A-neurons' second compartment. 
Querying the network consists in bringing some A-neurons to spike in response to the input pair $(inputOAi,\, inputMAii)$ and send backward input to O-neurons, so that these spike and send backward input to interface neurons (see Algorithm \ref{query}). The resulting set of firing interface neurons is the network's answer to the query. 

\begin{algorithm}[h!]
\caption{\query\ (Queries the neural network for an action's outcome)}\label{query}
\begin{algorithmic}[1]

\Input $(\inputOAi, \cnxOAi, \inputMAii, \cnxMAii, \cnxOAiii,  \cnxLO, \lastSpikedA)$
\LComment{Recall that: \\
    $\inputOAi \subset \O$ and $\inputMAii \subset \M$; \\
    $\cnxOAi$, $\cnxMAii$, $\cnxOAiii$ and $\cnxLO$ are matrices of integers of respective dimensions $(\card{\O} \times \card{\A})$, $(\card{\M} \times \card{\A})$, $(\card{\O \cup \F} \times \card{\A})$ and  \mbox{$(\card{\L} \times \card{\O})$;}  \\
    $\lastSpikedA$ is a vector of integers of length $\card{\A}$; \\
    $\TnbQueryA$ is a constant (the desired number of spiking A-neurons in the querying process).} 
\State $\inhibitedIneurons, \firedA\ \gets \emptyset$ \Comment{Subsets of, respectively, $\L \cup \F$ and \A}
\State $\backwardFiredI\ \gets \emptyset$  \Comment{dictionary with keys in $\L \cup \F$ and values in $\mathbb{R}$}

\State $\backwardInputSumOuF \gets [0, 0, ...~ , 0]$ \Comment{vector of integers of length $\card{\O \cup \F}$}
\State $\backwardInputSumL \gets [0, 0, ...~ , 0]$ \Comment{vector of integers of length  $\card{\L}$}
\State $\Fail\ \gets \mathrm{False}$ 

\State $\modulatedCnxOAi \gets$ \hyperref[modulateCnx]{\modulateCnx}$(\inputOAi, \cnxOAi)$
\State $\modulatedCnxMAii \gets$ \hyperref[modulateCnx]{\modulateCnx}$(\inputMAii, \cnxMAii)$
\State $\inputSumAi \gets$ \hyperref[calculateInputSumsA]{\calculateInputSumsA}$(\inputOAi, \modulatedCnxOAi)$
\State $\inputSumAii \gets$ \hyperref[calculateInputSumsA]{\calculateInputSumsA}$(\inputMAii, \modulatedCnxMAii)$
\State $\inputSumAiAii \gets$ \hyperref[sumInputsAiAii]{\sumInputsAiAii}$(\inputSumAi, \inputSumAii)$

\State $V \gets \{v \mid  (a, v) \in \inputSumAiAii\}$
\While{$\card{\firedA} <  \TnbQueryA$ and $V \neq \emptyset$ and $\Fail = \mathrm{False}$}
    \State $\currentAThreshold \gets  \max(V)$
    \State $\firedA \gets \{a \in \A \mid \inputSumAiAii[a]\geq  \currentAThreshold \}$
    \State $V \gets V \setminus\{\currentAThreshold\}$ 
    \State  $(\backwardInputSumOuF, \aboveBackwardSTOuF) \gets$ \hyperref[computeBackwardInputSumsO]{\computeBackwardInputSumsOuF}$(\firedA, \cnxOAiii,  \backwardInputSumOuF)$
    \State $\backwardFiredO \gets \emptyset$ \Comment{set of O-neurons that fired}
    \While{$\card{\backwardFiredO} < \TnbQueryO$ and $\aboveBackwardSTOuF \neq\emptyset$ and  $\Fail = \mathrm{False}$}
        \State $(\spikingO, \Fail, \backwardFiredI, \aboveBackwardSTOuF) \gets$ \hyperref[backwardSpikeO]{\backwardSpikeO}$(\aboveBackwardSTOuF, \inhibitedIneurons, \currentAThreshold, \backwardFiredI)$
        \State $\backwardFiredO \gets \backwardFiredO \cup \spikingO$
        \For{$o \in \spikingO$}
            \State $\backwardInputSumOuF[o] \gets 0$
        \EndFor
        \If{ $\Fail = \mathrm{False}$}
        
            \State $(\backwardFiredI, \backwardInputSumL, \inhibitedIneurons)
 \gets$ \hyperref[backwardSpikeIQuery]{\backwardSpikeIQuery}$(\spikingO, \cnxLO,  \backwardInputSumL, $ $\backwardFiredI, \inhibitedIneurons)$                                 
        \EndIf
    \EndWhile
\EndWhile
\Statex \Return  $(\backwardFiredI, \firedA)$

\end{algorithmic}
\end{algorithm}


Now, among the A-neurons that receive connections from $\inputOAi$ and $\inputMAii$ neurons, some support the representation of general action concepts (such as, e.g.,   \cpt{\cpt{OK},\,\cpt{NE},\,\cpt{OK}}), while others support the representation of more particular action concepts (such as, say,  \cpt{\cpt{OK, NorthWall},\,\cpt{NE},\linebreak\cpt{Failure}}). 
Obviously, it is desirable that the A-neurons supporting more particular concepts take precedence over those supporting more general concepts, as the information conveyed by more particular concepts is more accurate. 
To achieve this, the input sent to A-neurons is modulated by applying a multiplying factor $\coefMult$ to the connections' weights   (see Algorithm \ref{modulateCnx} in Appendices). 
More precisely, for any neuron $o$ in $\inputOAi$,  $\coefMult \in [1, 2]$  is a decreasing function of $\sum\cnxOAi_{a \in \A}[o][a]$, and similarly for  $\inputMAii$. 
Modulated input sums are then computed and combined, triggering the firing of a growing set of A-neurons (the process is the same as the one previously described in Algorithm \ref{spikeA}).
As O- and M- neurons with fewer connections to A-neurons tend to support more particular concepts than neurons with more connections, this modulation mechanism helps A-neurons encoding more particular actions to fire earlier and more repeatedly than A-neurons encoding more general actions. 

Each time an A-neuron spikes, it sends a backward input to O- and $\FailureNeuron$ neurons through its third compartment's connections. Backward input sums are computed incrementally as A-neurons fire (see Algorithm \ref{computeBackwardInputSumsO} in Appendices), bringing O- and $\FailureNeuron$ neurons to fire as soon as their spike threshold is reached (see Algorithm \ref{backwardSpikeO} in Appendices). If some O-neuron fires first, it inhibits the $\FailureNeuron$ neuron and prevents it from further firing. Similarly, if the $\FailureNeuron$ neuron spikes first, it inhibits all the O-neurons  and prevents them from further firing. 
Firing O-neurons in turn send backward inputs to L-neurons, and the input sum received by L-neurons is computed incrementally. L-neurons spike as soon as they reach their threshold $\STL$ (see Algorithm \ref{backwardSpikeIQuery} in Appendices), and their firing inhibits the L-neurons that encode incompatible features (for instance, the firing of the L-neuron encoding the feature \OKFname\  prevents the L-neuron encoding the feature \KOFname\ from further firing).

Due to input modulation, the input sum received by a given A-neuron $a$ is a good indicator of both the accuracy of the action concept encoded by $a$ relative to the input pair $(\inputOAi, \inputMAii)$, and of the concept's degree of generality: the more adequate and particular the encoded concept, the higher this input sum. 
The value of A-neurons' spike threshold at the moment a given neuron  $i \in  \L \cup \F$ spikes can thus be used to assess the relevance of the A-neurons involved in $i$'s spiking relative to the input pair $(\inputOAi, \inputMAii)$.
This provides the agent with a means to estimate the reliability of its prediction of the feature $f_i$ encoded by $i$: the higher A-neurons' spike threshold at the moment $i$ spiked, the higher this reliability. 
To keep track of this information, the set $\backwardFiredI$ returned by  Algorithm \ref{backwardSpikeIQuery} and further returned by the querying process is a set of pairs $(i, c)$, where $c$ is A-neurons' spike threshold at the moment $i$ spiked.

\subsection{Making Decisions}\label{sec:MakingDecisions}

Suppose that the agent is at some location and wants to make a step. The process by which it decides which motor activity to perform (i.e., in which direction to go) is run by its decision system 
(see Algorithm \ref{chooseAMove}).
%
%
%
\begin{algorithm}[t]
\caption{\chooseAMove\ (decides the agent's next move)}\label{chooseAMove}
\begin{algorithmic}[1]
\Input $(\inputOAi, \cnxOAi, \cnxMAii, \cnxOAiii, \cnxLO, \lastSpikedA)$
\LComment{Recall that: \\
    $\inputOAi \subset \O$; \\
    $\cnxOAi$, $\cnxMAii$, $\cnxOAiii$ and $\cnxLO$ are matrices of integers of respective dimensions \mbox{$(\card{\O} \times \card{\A})$, $(\card{\M} \times \card{\A})$, $(\card{\O \cup \F} \times \card{\A})$ and  $(\card{\L} \times \card{\O})$;}  \\
    $\lastSpikedA$ is a vector of integers of length $\card{\A}$.} 
\State $\selectedMode \gets$ \hyperref[randomChoice]{\randomChoice}$(\{$\emph{``Exploration''},\emph{``Exploitation''}$\})$
\State $\predictions \gets$  \hyperref[makePredictions]{\makePredictions} $(\inputOAi, \cnxOAi, \cnxMAii, \cnxOAiii, \cnxLO, \lastSpikedA)$
\State $(\suitable, \unsuitable, \undecided)  \gets$  \hyperref[ratePredictions]{\ratePredictions}$(\predictions)$
\State $\move \gets$ \hyperref[makeAChoice]{\makeAChoice}$(\suitable, \unsuitable, \undecided, \selectedMode)$
\State  $\predictedFeatures \gets  \{f \mid (f,  \confi{f}) \in \predictions[\move]\}$ 
 \Statex \Return $(\move, \predictedFeatures)$

\end{algorithmic}
\end{algorithm}
First, the agent decides whether to exploit its current knowledge about its environment, or to explore its environment to improve its knowledge. The exploration/exploitation dilemma is a well-known problem in online learning \cite{Wat89,SutBar18}, and changing environments make it even more difficult. For this reason we do not try to reach an optimal solution here, but we simply make the agent's decision system choose at random with equal probability between an \emph{Exploration} and an \emph{Exploitation} mode.  

The agent then makes predictions about the outcome of each possible action. To do so, it successively queries its semantic memory for the outcome of the action having the current location as initial situation and one of the eight possible moves as motor activity  (see  Algorithm \ref{makePredictions}).
%
%
\begin{algorithm}[t]
\caption{\makePredictions\ (makes predictions about the outcome of each motor activity given the current situation)}\label{makePredictions}
\begin{algorithmic}[1]
\Input $(\inputOAi, \cnxOAi, \cnxMAii, \cnxOAiii, \cnxLO, \lastSpikedA)$
\LComment{Recall that: \\
    $\inputOAi \subset \O$; \\
    $\cnxOAi$, $\cnxMAii$, $\cnxOAiii$ and $\cnxLO$ are matrices of integers of respective dimensions $(\card{\O} \times \card{\A})$, $(\card{\M} \times \card{\A})$, $(\card{\O \cup \F} \times \card{\A})$ and  \mbox{$(\card{\L} \times \card{\O})$;}\\  
    $\lastSpikedA$ is a vector of integers of length $\card{\A}$; \\
    $\MotAct$ is the set of the agent's possible motor activities and
    $\LF$ the set of locations' features; \\
    $\backwardFiredI$ is a dictionary whit keys in $\L \cup \F$ and values in $\mathbb{R}$.} 
\State $\predictions \gets \emptyset$ \Comment{dictionary with keys in $\MotAct$ and values which are themselves dictionaries with keys in $\LF\cup \{\FailFname\}$ and values in $\mathbb{R}$.}
\For{$\move \in \MotAct$}
    \State $m \gets$ index of the M-neuron encoding the motor activity feature corresponding to $\move$.
    \State $m' \gets$ index of the M-neuron encoding the motor activity feature \DiagFname\ or \OrthFname, depending on whether $\move$ is a diagonal or orthogonal move.
    \State $\inputMAii \gets \{m,\, m'\}.$ 
    \State $(\backwardFiredI, \firedA) \gets$ \hyperref[query]{\query}$(\inputOAi, \cnxOAi, \inputMAii,  \cnxMAii, $ $\cnxOAiii, \cnxLO, \lastSpikedA)$
    \State $\predictions[\move] \gets \emptyset$ \Comment{\mbox{dictionary with keys in $\LF \cup \{\FailFname\}$ and values in $\mathbb{R}$}}
    \If{$\backwardFiredI \neq \emptyset$} 
        \State $\shortTermMemoryCoef  \gets 1 + (1/\mean(\{\lastSpikedA[a] \mid a \in \firedA\}))$     
        \For{$i \in \{i \mid (i, \mathit{threshold}) \in \backwardFiredI\}$ } 
            \State $f\gets$ feature encoded by $i$
            \State $\confi{f}\gets \backwardFiredI[i]* \shortTermMemoryCoef$
            \State $\predictions[\move][f] \gets \confi{f}$
        \EndFor         
    \EndIf
 \EndFor
\Statex \Return $\predictions$
\end{algorithmic}
\end{algorithm}
After each such querying process, a coefficient $\shortTermMemoryCoef$ is computed from the $\lastSpikedA$ values of the A-neurons firing in the course of the query process. $\shortTermMemoryCoef = 2$ if all the A-neurons firing in the querying process learned at the previous step ---which can only occur if the queried action is the same as the action performed at the previous step, i.e., if the action's outcome is a failure--- and quickly decreases as more steps were performed since the neurons' last learning episode to get close to 1.
$\shortTermMemoryCoef$ represents the agent's working memory, that is, its memory of having just performed the considered action. 
For each interface neuron $i$ firing as a result of the querying process, the feature $f$ encoded by $i$ is then retrieved, and a confidence degree $\confi{f}$ is computed from $\shortTermMemoryCoef$  and the A-neurons' spike threshold at the moment $i$ fired. 
$\confi{f}$ represents the confidence the agent has in its prediction of $f$. It corresponds to the agent's estimation of how well the action concepts on which it relied to make its prediction fit the envisaged action, weighted by its short term memory.

The prediction process thus returns a set (dictionary)
\begin{align*}
\predictions &= \{(\move,\, \predictions[\move]) \mid \move \in \MotAct\},
\intertext{where}
\predictions[\move] &= \{(f_1, \confi{f_1}), ... (f_n, \confi{f_n})\}
\end{align*}
is itself a dictionary which to each feature $f_i$ predicted by the agent associates the degree of confidence $\confi{fi}$ the agent has in its prediction of $f_i$.

The agent then rates each motor activity $\move$ for its suitability  (see  Algorithm \ref{ratePredictions} in Appendices), by building the sets $\suitable$,  $\unsuitable$ and  $\undecided$ such that: 
\begin{align*}
\suitable & = \{(\move,\, \confi{OK}) \mid (\OKFname, \confi{OK}) \in \predictions[\move] \}\\
\unsuitable & \{(\move, \confi{f}) \mid (f = \KOFname \textrm{ or } f = \FailFname) \textrm{ and } (f, \, \confi{f}) \in \predictions[\move] \}\\
\undecided & = \{\move\mid  \nexists \conf \textrm{ such that } (\move,\conf) \in \suitable \cup \unsuitable\}
\end{align*}

Finally, the decision system chooses a motor activity depending on the selected mode (see  Algorithm \ref{makeAChoice} in Appendices). In \emph{Exploration} mode, the agent is willing to take risks and chooses an action with the most uncertain outcome possible so to increase its knowledge: if $\undecided \neq \emptyset$ it picks one from it, otherwise it goes for one with the least $\conf$ in $\suitable\cup \unsuitable$.  
In \emph{Exploitation} mode  by contrast, the agent just wants to avoid KO boxes and failure as much as possible. So, if $\suitable\neq\emptyset$ it chooses a move with the greatest $\conf$, otherwise if $\undecided\neq\emptyset$ it picks one from it, and if both $\suitable$ and $\undecided$ are empty it chooses a move with the least $\conf$ in $\unsuitable$.

\subsection{Making a step}\label{sec:makeAStep}

Let $\departLoc$ be the agent's current location, and $\prevFiredO$ be the set of firing O-neurons resulting from the agent considering that location after having learned about it at the previous step ($\prevFiredO$ is initialized to $\emptyset$ for the agent's first step). To make a step, the agent proceeds as follows (see Algorithm \ref{makeAStep}).
%
\begin{algorithm}[t!]
\caption{\makeAStep\ (specifies how the agent makes a step)}\label{makeAStep}
\begin{algorithmic}[1]

\Input $(\departLoc, \prevFiredO, \cnxLO, \cnxOAi, \cnxMAii, \cnxOAiii, \lastSpikedO, \lastSpikedA)$
\LComment{Recall that: \\
    $\prevFiredO \subset \O;$\\
    $\cnxOAi$, $\cnxMAii$, $\cnxOAiii$ and $\cnxLO$ are matrices of integers of respective dimensions \mbox{$(\card{\O} \times \card{\A})$, $(\card{\M} \times \card{\A})$, $(\card{\O \cup \F} \times \card{\A})$ and $(\card{\L} \times \card{\O})$;}\\  
    $\lastSpikedO$ and $\lastSpikedA$ are vectors of integers of respective lengths $\card{\O}$ and $\card{\A}$;}
\State $\inputOAi\ \gets \prevFiredO$
\State $(\move, \predictedFeatures) \gets$  \hyperref[chooseAMove]{\chooseAMove}$(\inputOAi, \cnxOAi, \cnxMAii,$ $ \cnxOAiii, \cnxLO, \lastSpikedA)$
\State $\inputMAii \gets  \{m, m'\} $ such that $m$ is the M-neuron encoding the motor activity $\move$ and $m'$ is the M-neuron encoding \DiagFname\ or \OrthFname, depending on the case.

\State $(\newLoc, \newLocFeatures) \gets$ \hyperlink{calculateNewLoc}{\calculateNewLoc}$(\departLoc, \move)$
\If{$\newLoc = \departLoc$}
    \State $\firedO \gets \prevFiredO$ \Comment{(no need to compute $\firedO$ again)}
    \State $\inputOAiii \gets$ \F 
    \State $\actResultIneurons \gets$ \F 
\Else
    \State $\inputLO \gets  \{l \in \L \mid \exists f \in \newLocFeatures$ such that $l$ encodes $f\}$
    \State $(\inputSumO, \firedO) \gets$  \hyperref[spikeO]{\spikeO}$(\inputLO, \cnxLO)$
    \State $\cnxLO \gets$ \hyperref[learnO]{\learnO}$(\inputLO, \inputSumO, \firedO, \cnxLO, \lastSpikedO)$
    \State $(\inputSumO, \firedO) \gets$ \hyperref[spikeO]{\spikeO}$(\inputLO, \cnxLO)$
    \State $\inputOAiii \gets \firedO$
    \State $\actResultIneurons \gets \inputLO$
\EndIf  
\State $\predFeatIneurons \gets \{n \in \L \cup \F \mid \exists f \in \predictedFeatures$ and $n$ encodes $f\}$
\State $(\cnxOAi, \cnxMAii, \cnxOAiii) \gets$ \hyperref[learnA]{\learnA}$(\inputOAi, \cnxOAi, \inputMAii, \cnxMAii, \inputOAiii, $ $ \cnxOAiii, \actResultIneurons, \predFeatIneurons, \lastSpikedA)$
\State  $\modulatedCnxOAi \gets$ \hyperref[modulateCnx]{\modulateCnx}$(\inputOAi, \cnxOAi)$
\State $\modulatedCnxMAii \gets$ \hyperref[modulateCnx]{\modulateCnx}$(\inputMAii, \cnxMAii)$
\State $(\inputSumAiAii, \firedA)\gets$  \hyperref[spikeA]{\spikeA}$(\inputOAi, \inputMAii, \modulatedCnxOAi, \modulatedCnxMAii)$

\State $\departLoc\  \gets \newLoc$
\State $\prevFiredO\  \gets \firedO$

\State $\lastSpikedO \gets$ \hyperref[updateLastSpikes]{\updateLastSpikes}$(\lastSpikedO, \firedO)$
\State $\lastSpikedA \gets$ \hyperref[updateLastSpikes]{\updateLastSpikes}$(\lastSpikedA, \firedA)$

\Statex \Return  $(\departLoc, \prevFiredO, \cnxLO, \cnxOAi, \cnxMAii, \cnxOAiii, \lastSpikedO, \lastSpikedA)$ 
\end{algorithmic}
\end{algorithm}

First, it decides in which direction to go by running the decision process described in Section \ref{sec:MakingDecisions}. $\prevFiredO$ is the input to A-neurons' first compartment. 
Once the agent's decision is made, the decision system transmits the information to the motor system which performs the selected motor activity. The M-neurons encoding the motor activity's features are activated by proprioception and send an input $\inputMAii$ to A-neurons' second compartment.

The agent's move is simulated by a function \hypertarget{calculateNewLoc}{$\calculateNewLoc$}, which computes the agent's arrival location and the set of its features. If the arrival location is the same as the initial one (that is, if the agent bumped into a wall) the $\FailureNeuron$ neuron fires and sends an input $\inputOAiii$ to A-neurons' third compartment. The agent then directly proceeds to learn the action (see A-neurons' learning in Section \ref{sec: A-neurons}).
Otherwise, it first observes its arrival location and learns about it. Do do so, the agent's perceptual system triggers the firing of the set $\inputLO$ of L-neurons that encode the location's features, which in turn triggers the firing of a set $\firedO$ of O-neurons. 
These are the neurons that encode the agent's concepts that best fit the observed situation and that are therefore the best suited for updating. 
The object concepts corresponding to the arrival location are then learned according to the learning process described in Section \ref{sec: O-neurons}.
This being done, the agent considers once again its arrival location: interface neurons from $\inputLO$ fire again, triggering the firing of a new set $\firedO$ of O-neurons (which may, or not, be the same as previously). These neurons encode the agent's updated knowledge about its arrival location. They are used as input set $\inputOAiii$ to A-neurons' third compartment. 
The agent then proceeds to learn the action, according to the learning process described in Algorithm \ref{learnA}.

Once the action is learned, the input pair $(\inputOAi, \inputMAii)$ is sent again to A-neurons' first two compartments, using connections' modulation as in the querying process (see Section \ref{sec:Querying}). The neurons that fire in response to these inputs are the ones that encode the agent's updated knowledge about the action. 

The agent is now ready for the next step. The current arrival location is next step's depart location, and  the current set $\firedO$ is next step's $\prevFiredO$. O- and A-neurons' last spikes values are updated: the neurons that did not spike after learning have their values increased by one while those that did spike have their values set to 1 (reset to zero and then increased by one for the next step).

\section{Experiments and results}\label{sec: Results}

The testing of the agent's abilities was carried out by placing it at location $(0,0)$ and prompting it to perform a succession of series of steps, each complete sequence of series of steps being called a \emph{trial}. 
Trials are therefore sequences  $t = [s_1, ... , s_{n-1}, s_n]$, where each series of steps $s_i$ is a number of steps to be performed;  the total number of steps in the trial is $\sum_{i \in \{1, ... , n\}}\hspace{-.2em}s_i$. 
For brevity, in plots and tables series of steps are denoted by the total number of steps performed since the beginning of the trial at the moment of their last step. This means that the series $s_j$ is denoted by the integer $n = \sum_{i \leq  j}\hspace{-.1em}s_i$.

The results presented here are averaged over 50 trials. Three distinct groups of tests (\emph{``Experiments''}) were carried out. The code provided in supplementary material allows to reproduce the experiments while varying the number of trials, sequences, steps in sequences, and others parameters.
%

\subsection{Experiment \#1}
In a first group of tests, trials were sequences 
\begin{center}
 $t = [1, 1, 2, 4, 8, 16, 32, 64, 128, 256, 512, 1024, 2048, 4096, 8192, 16384, 32768]$,
\end{center}
which is 65536 steps in total. The door was kept closed all along, so the agent had no access to the second room. 

A first test concerned the agent's ability to learn an action over one single experience (``one-shot learning''). To assess it, after each step  the agent was asked to redo the prediction that led to the just realized action, and this new prediction was  compared with the action's actual outcome (see Table \ref{tab: Post-Learning percentages}). 
\begin{table}[t]
\centering
\caption{Post-learning predictions percentages (mean over 50 trials). CC: Correct and Complete predictions, MF: Missed Features, PE: Prediction Errors.}
\label{tab: Post-Learning percentages} 
 \setlength{\tabcolsep}{2.1mm}
\fontsize{9pt}{9pt}\selectfont
\begin{tabular}{c| c  c  c  c  c  c  c  c  c } 
\toprule
\parbox{1.2cm}{Total nb\\ of Steps}& 1& 2& 4& 8& 16& 32& 64& 128& 256\\
 \midrule  
CC & 0.0 & 100.0 & 99.0 & 99.5 & 99.8 & 99.0 & 97.3 & 91.8 & 88.2\\
MF & 100.0 & 0.0 & 0.3 & 0.2 & 0.1 & 0.6 & 1.7 & 4.6 & 6.1\\
PE & 0.0 & 0.0 & 0.0 & 0.0 & 0.0 & 0.0 & 0.1 & 0.1 & 0.3\\
\midrule
\midrule
\parbox{1.2cm}{Total nb\\ of Steps}& 512& 1024& 2048& 4096& 8192& 16384& 32768& 65536& \multicolumn{1}{ |c}{Global}\\
\midrule 
CC & 90.0 & 93.2 & 95.1 & 95.6 & 96.1 & 96.3 & 97.1 & 97.0 & \multicolumn{1}{ |c}{96.7}\\
MF & 5.0 & 3.2 & 2.1 & 2.0 & 1.6 & 1.6 & 1.2 & 1.3 & \multicolumn{1}{ |c}{1.4}\\
PE & 0.3 & 0.2 & 0.2 & 0.1 & 0.1 & 0.1 & 0.1 & 0.1 & \multicolumn{1}{ |c}{0.1}\\
\bottomrule 
\end{tabular}
\end{table}
A prediction is said to be \emph{Correct and Complete} (CC) if the predicted features are exactly those of the arrival location. 
The table's first line shows the percentage of performed steps leading to a CC post-learning prediction for each series of steps.  

At each step, the arrival location's features (\emph{``features to predict''})  were listed and counted, and those among them that the agent failed to predict (\emph{``Missed Features''}) were counted. The table's second line (MF for \emph{``Missed Features''}) shows the percentage of features to predict that the agent failed to predict for each series of steps. 

The features predicted by the agent were also listed and counted at each step, and those among them that were wrongly predicted (i.e., that did not belong to the arrival location's feature set) were counted.
The table's third line (PE for \emph{``Predictions Errors''}) shows the percentage of predicted features that were wrongly predicted  for each series of steps. 

These results show a good performance at immediate recall after learning. 
The fact that the network does not learn anything at the first step is expected, as there is no previous step hence $\inputOAi$ is the empty set. This means that A-neurons receive no input on their first compartment, so the inputs received on their second compartment cannot be integrated. It follows that the total input received by A-neurons at the first step is zero, hence no learning can occur.  
 The lower scores around 128-1024 steps are explained by the fact that at start-up the network has a lot of unused neurons at its disposal and simply learns the particular concepts with all their features, so specific features are well learned. As the network accumulates knowledge, unused neurons get scarce and the network tends to recruit less-used neurons for learning instead. A trade-off is then made between learning the current observation’s specific features and not forgetting the common features already encoded by the neurons,  which makes specific features more difficult to learn. Yet after repeated encounters the specific features are finally learned, so the scores improve again.

\medskip

To test whether the acquired knowledge was retained in the long run, after each series of steps the simulation was frozen and learning  was deactivated, and the agent was placed successively in each location of each room. There, it was queried for its predictions for each of the eight possible motor activities. Its predictions were recorded and compared with the actions' actual outcomes. 
Tables \ref{tab: Hit Rates Door Closed} and \ref{tab: Correctness Door Closed} show each feature's mean \emph{Hit Rate} (that is, its chances of being predicted when effectively present), and \emph{Correctness} (its chances of being effectively present when predicted)\footnote{\emph{Hit Rate} is also known as \emph{True Positive Rate}, \emph{Recall} or \emph{Sensitivity}, while \emph{Correctness} is also known as \emph{Precision} or \emph{Positive Predictive Value}--- see for example \cite{KohEtPro98} for definitions. Here we multiplied the obtained figures by 100 to get percentages.}  for each room. 


\begin{table}[t]
\caption{Predictions' \emph{Hit Rates} after $n$ steps (mean over 50 trials).} 
\label{tab: Hit Rates Door Closed} 
\centering
\setlength{\tabcolsep}{2mm}
 \renewcommand{\arraystretch}{1.1}
\fontsize{9pt}{9pt}\selectfont 
\begin{tabular}{c|r| r r r r r r r } 
\toprule\rotatebox{90}{\parbox{.8cm}{Nb of\\ Steps}} &\rotatebox{90}{Room} & \rotatebox{90}{OK}& \rotatebox{90}{KO} & \rotatebox{90}{Fail.} & \rotatebox{90}{Wall} & \rotatebox{90}{Cold}& \rotatebox{90}{Sound} & \rotatebox{90}{\parbox{.8cm}{Box \\ Name}}\\
\hline
\rule{-2pt}{2.2ex}
 \multirow{2}{*}{2} & 1  & 10.0 & 10.0 & 0.0 & 2.6 & 2.9 & 0.0 & 1.1\\
&\cellcolor{black!15}2 & \cellcolor{black!15} 7.6 & \cellcolor{black!15} 9.1 & \cellcolor{black!15} 0.0 & \cellcolor{black!15} 1.6 & \cellcolor{black!15} 0.0 & \cellcolor{black!15} 0.0 & \cellcolor{black!15} 0.0\\
 \multirow{2}{*}{8} & 1  & 27.8 & 28.7 & 14.1 & 8.7 & 6.8 & 0.0 & 4.5\\
&\cellcolor{black!15}2 & \cellcolor{black!15} 19.5 & \cellcolor{black!15} 21.8 & \cellcolor{black!15} 12.4 & \cellcolor{black!15} 5.0 & \cellcolor{black!15} 0.0 & \cellcolor{black!15} 0.0 & \cellcolor{black!15} 0.0\\
 \multirow{2}{*}{64} & 1  & 87.9 & 85.3 & 39.7 & 36.9 & 30.0 & 0.0 & 22.4\\
&\cellcolor{black!15}2 & \cellcolor{black!15} 78.5 & \cellcolor{black!15} 81.4 & \cellcolor{black!15} 32.3 & \cellcolor{black!15} 23.6 & \cellcolor{black!15} 0.0 & \cellcolor{black!15} 0.0 & \cellcolor{black!15} 0.0\\
 \multirow{2}{*}{512} & 1  & 94.1 & 90.7 & 69.6 & 63.1 & 55.2 & 0.0 & 49.4\\
&\cellcolor{black!15}2 & \cellcolor{black!15} 86.2 & \cellcolor{black!15} 83.0 & \cellcolor{black!15} 41.1 & \cellcolor{black!15} 21.3 & \cellcolor{black!15} 0.0 & \cellcolor{black!15} 0.0 & \cellcolor{black!15} 0.0\\
 \multirow{2}{*}{4096} & 1  & 97.2 & 94.1 & 84.2 & 83.2 & 77.8 & 0.0 & 73.1\\
&\cellcolor{black!15}2 & \cellcolor{black!15} 89.2 & \cellcolor{black!15} 82.7 & \cellcolor{black!15} 54.0 & \cellcolor{black!15} 27.7 & \cellcolor{black!15} 0.0 & \cellcolor{black!15} 0.0 & \cellcolor{black!15} 0.0\\
 \multirow{2}{*}{16384} & 1  & 98.4 & 96.5 & 88.8 & 87.6 & 87.7 & 0.0 & 81.8\\
&\cellcolor{black!15}2 & \cellcolor{black!15} 90.0 & \cellcolor{black!15} 83.7 & \cellcolor{black!15} 56.9 & \cellcolor{black!15} 29.1 & \cellcolor{black!15} 0.0 & \cellcolor{black!15} 0.0 & \cellcolor{black!15} 0.0\\
 \multirow{2}{*}{32768} & 1  & 98.4 & 96.3 & 91.0 & 88.6 & 84.0 & 0.0 & 83.6\\
&\cellcolor{black!15}2 & \cellcolor{black!15} 89.5 & \cellcolor{black!15} 84.0 & \cellcolor{black!15} 59.6 & \cellcolor{black!15} 27.6 & \cellcolor{black!15} 0.0 & \cellcolor{black!15} 0.0 & \cellcolor{black!15} 0.0\\
 \multirow{2}{*}{65536} & 1  & 98.5 & 96.9 & 91.1 & 89.8 & 87.8 & 0.0 & 84.5\\
&\cellcolor{black!15}2 & \cellcolor{black!15} 92.9 & \cellcolor{black!15} 84.2 & \cellcolor{black!15} 59.6 & \cellcolor{black!15} 29.7 & \cellcolor{black!15} 0.0 & \cellcolor{black!15} 0.0 & \cellcolor{black!15} 0.0\\
\bottomrule 
\end{tabular}
 \end{table}


\begin{table}[t]
\caption{Predictions' \emph{Correctness} after $n$ steps (mean over 50 trials).} 
\label{tab: Correctness Door Closed} 
\centering
\setlength{\tabcolsep}{2mm}
\renewcommand{\arraystretch}{1.1}
\fontsize{9pt}{9pt}\selectfont 
\begin{tabular}{c|r| r r r r r r r } 
\toprule\rotatebox{90}{\parbox{.8cm}{Nb of\\ Steps}} &\rotatebox{90}{Room} & \rotatebox{90}{OK}& \rotatebox{90}{KO} & \rotatebox{90}{Fail.} & \rotatebox{90}{Wall} & \rotatebox{90}{Cold}& \rotatebox{90}{Sound} & \rotatebox{90}{\parbox{.8cm}{Box \\ Name}}\\
\hline
\rule{-2pt}{2.2ex}
 \multirow{2}{*}{2} & 1  & 58.8 & 31.0 & 0.0 & 19.7 & 4.3 & 0.0 & 18.4\\
&\cellcolor{black!15}2 & \cellcolor{black!15} 12.0 & \cellcolor{black!15} 14.2 & \cellcolor{black!15} 0.0 & \cellcolor{black!15} 2.0 & \cellcolor{black!15} 0.0 & \cellcolor{black!15} 0.0 & \cellcolor{black!15} 0.0\\
 \multirow{2}{*}{8} & 1  & 79.1 & 73.6 & 40.4 & 29.2 & 7.1 & 0.0 & 15.3\\
&\cellcolor{black!15}2 & \cellcolor{black!15} 53.8 & \cellcolor{black!15} 49.0 & \cellcolor{black!15} 31.3 & \cellcolor{black!15} 10.4 & \cellcolor{black!15} 0.0 & \cellcolor{black!15} 0.0 & \cellcolor{black!15} 0.0\\
 \multirow{2}{*}{64} & 1  & 81.1 & 77.4 & 67.6 & 33.2 & 25.6 & 0.0 & 22.1\\
&\cellcolor{black!15}2 & \cellcolor{black!15} 81.3 & \cellcolor{black!15} 79.6 & \cellcolor{black!15} 51.9 & \cellcolor{black!15} 17.6 & \cellcolor{black!15} 0.0 & \cellcolor{black!15} 0.0 & \cellcolor{black!15} 0.0\\
 \multirow{2}{*}{512} & 1  & 89.8 & 86.9 & 80.9 & 73.4 & 77.0 & 0.0 & 69.0\\
&\cellcolor{black!15}2 & \cellcolor{black!15} 83.6 & \cellcolor{black!15} 83.1 & \cellcolor{black!15} 48.7 & \cellcolor{black!15} 28.0 & \cellcolor{black!15} 0.0 & \cellcolor{black!15} 0.0 & \cellcolor{black!15} 0.0\\
 \multirow{2}{*}{4096} & 1  & 93.8 & 93.1 & 91.7 & 88.1 & 92.7 & 0.0 & 85.4\\
&\cellcolor{black!15}2 & \cellcolor{black!15} 86.1 & \cellcolor{black!15} 86.1 & \cellcolor{black!15} 60.6 & \cellcolor{black!15} 37.3 & \cellcolor{black!15} 0.0 & \cellcolor{black!15} 0.0 & \cellcolor{black!15} 0.0\\
 \multirow{2}{*}{16384} & 1  & 95.3 & 95.1 & 95.9 & 91.5 & 94.7 & 0.0 & 90.2\\
&\cellcolor{black!15}2 & \cellcolor{black!15} 86.4 & \cellcolor{black!15} 87.3 & \cellcolor{black!15} 62.6 & \cellcolor{black!15} 39.6 & \cellcolor{black!15} 0.0 & \cellcolor{black!15} 0.0 & \cellcolor{black!15} 0.0\\
 \multirow{2}{*}{32768} & 1  & 95.8 & 96.4 & 95.6 & 93.2 & 95.5 & 0.0 & 91.8\\
&\cellcolor{black!15}2 & \cellcolor{black!15} 86.6 & \cellcolor{black!15} 88.3 & \cellcolor{black!15} 63.9 & \cellcolor{black!15} 39.9 & \cellcolor{black!15} 0.0 & \cellcolor{black!15} 0.0 & \cellcolor{black!15} 0.0\\
 \multirow{2}{*}{65536} & 1  & 95.9 & 96.3 & 96.6 & 92.9 & 94.9 & 0.0 & 91.9\\
&\cellcolor{black!15}2 & \cellcolor{black!15} 86.8 & \cellcolor{black!15} 88.7 & \cellcolor{black!15} 66.9 & \cellcolor{black!15} 38.9 & \cellcolor{black!15} 0.0 & \cellcolor{black!15} 0.0 & \cellcolor{black!15} 0.0\\
\bottomrule 
\end{tabular}
 \end{table}

Values for the first room (white lines) show that learned actions are indeed recalled long after having been performed. Values for the second room (grey lines) show that despite never having been in this room (since the door was kept closed) the agent was able to correctly predict \OKFname\ and \KOFname\ features and to a lesser extent \FailFname\ ---and this, even though locations from the second room have different sets of features, including for some of them a new feature, \SoundFname. 
The poor performance at wall prediction is due to the lack of general rules of the universe observable in the first room regarding the presence of walls in adjacent boxes. The agent thus relies on concepts of particular actions (i.e., action concepts involving particular concepts of individual boxes) to predict walls in the first room, but these cannot be reliably used in the second room.
The mixed result at failure prediction comes from a competition between general action concepts, the control of which needs to be improved.

\medskip

\begin{figure*}[t!]
    \centering
    \begin{subfigure}[b]{0.5\textwidth}
        \centering
        \includegraphics[width=.92\textwidth]{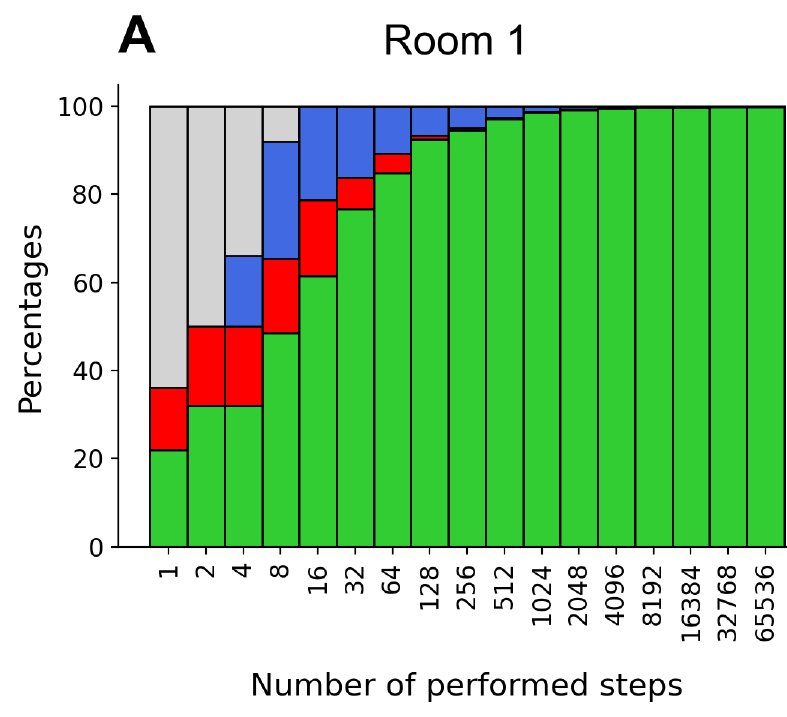}
    \end{subfigure}%
    ~ 
    \begin{subfigure}[b]{0.5\textwidth}
        \centering
        \includegraphics[width=.92\textwidth]{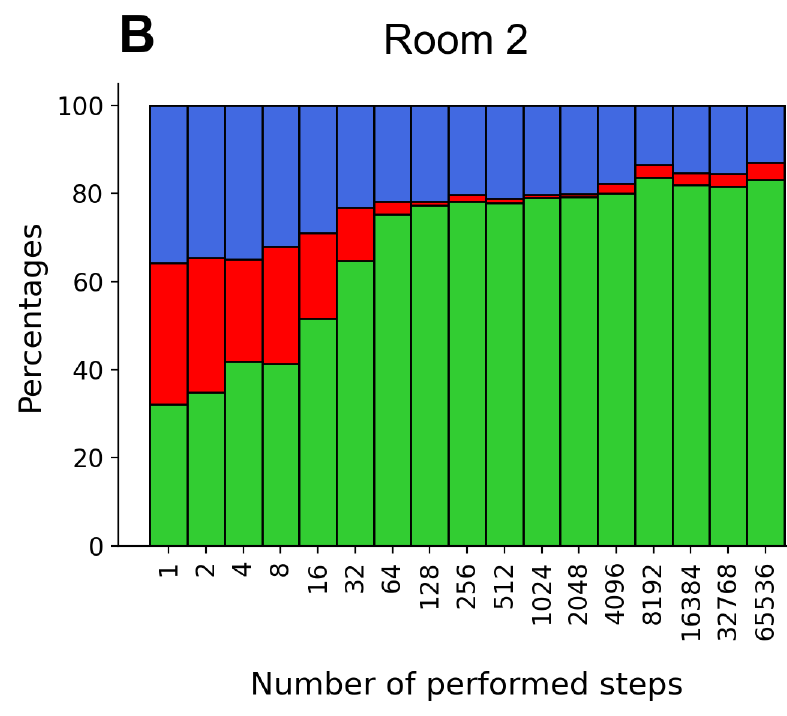}
    \end{subfigure}
    \caption{Actions' mean outcomes with door closed; Green: OK, Red: KO, Blue: Failure, Grey: no data.}
    \label{fig:Pictures/OK ResultsR1R2}
\end{figure*}

To test the agent's ability to use its knowledge to make appropriate decisions, the outcome of each action taken in \textit{Exploitation} mode was recorded throughout the trials. Figure~\ref{fig:Pictures/OK ResultsR1R2}.A shows the percentages of \OKFname, \KOFname, and \FailFname\ outcomes obtained in this manner for each series of steps (``no data'' corresponds to trials for which no steps were taken in \emph{Exploitation} mode in the considered series).
Visited locations were logged to check that the agent was not looping indefinitely on the same boxes: all boxes kept being visited, be it very rarely, at any point of the trials, due to the \emph{Exploration} mode. 
In average, most visited boxes were visited in about 10\% of the taken steps, versus 2\% for the least visited ones. Most visited locations are the OK boxes with no walls, which is expected since they are visited in both \emph{Exploration} and \emph{Exploitation} modes and can be accessed from all directions. Least visited locations are KO boxes with walls, which are less accessible and are mostly visited in \emph{Exploration} mode.
\medskip

Finally, the agent's ability to use the knowledge acquired in the first room to act judiciously in the second room was tested. At the end of each series of steps the agent was asked to chose a move from each of the second room's type of locations (where two locations are said to be of the same type if they have exactly the same features). 
Figure~\ref{fig:Pictures/OK ResultsR1R2}.B shows the percentages of \OKFname, \KOFname, and \FailFname\ outcomes obtained in this manner. These results reflect the agent's performance at making predictions about the second room's locations: it successfully predicts OK and KO boxes, but has more difficulties predicting failure.
%

\subsection{Experiment \#2}
To test the agent's ability to handle changes, a second group of tests was carried out. The  setup was the same as in Experiment \#1, except that the door was opened at the 2048$^{th}$ step  (as in Figure \ref{fig: Universe}.B). The agent spontaneously went in the second room, and spent a variable but significant amount of time in it (38.3\% of steps on average, standard deviation = 6.1). 
Tables \ref{tab: Hit Rates Door Opened} and \ref{tab: Correctness Door Opened} show the features' \emph{Hit Rates} and \emph{Correctness} from the moment the door was opened. 
These results show that the agent was able to learn new concepts involving the \SoundFname\ feature. The rather low \emph{Hit Rates} for the feature are due to the lack of observable cues in boxes at the direct south of boxes with sound, which prevents the agent from being able to predict it in 40\% of the cases.
\begin{table}[t]
\centering
 \caption{Predictions' \emph{Hit Rates} after $n$ steps, with the door opened at the 2048$^{th}$ step.} 
 \label{tab: Hit Rates Door Opened}
\setlength{\tabcolsep}{2mm}
 \renewcommand{\arraystretch}{1.1}
\fontsize{9pt}{9pt}\selectfont 
\begin{tabular}{c|r| r r r r r r r } 
\toprule\rotatebox{90}{\parbox{.8cm}{Nb of\\ Steps}} &\rotatebox{90}{Room} & \rotatebox{90}{OK}& \rotatebox{90}{KO} & \rotatebox{90}{Fail.} & \rotatebox{90}{Wall} & \rotatebox{90}{Cold}& \rotatebox{90}{Sound} & \rotatebox{90}{\parbox{.8cm}{Box \\ Name}}\\
\hline
\rule{-2pt}{2.2ex}
 \multirow{2}{*}{2048} & 1  & 96.4 & 94.0 & 79.8 & 78.0 & 73.8 & 0.0 & 68.2\\
&\cellcolor{black!15}2 & \cellcolor{black!15} 86.5 & \cellcolor{black!15} 83.3 & \cellcolor{black!15} 46.4 & \cellcolor{black!15} 25.5 & \cellcolor{black!15} 0.0 & \cellcolor{black!15} 0.0 & \cellcolor{black!15} 0.0\\
 \multirow{2}{*}{4096} & 1  & 92.9 & 94.2 & 80.0 & 61.8 & 57.8 & 0.0 & 51.7\\
&\cellcolor{black!15}2 & \cellcolor{black!15} 85.8 & \cellcolor{black!15} 90.0 & \cellcolor{black!15} 60.8 & \cellcolor{black!15} 35.3 & \cellcolor{black!15} 0.0 & \cellcolor{black!15} 34.3 & \cellcolor{black!15} 2.4\\
 \multirow{2}{*}{8192} & 1  & 94.8 & 95.6 & 79.6 & 66.5 & 61.8 & 0.0 & 57.9\\
&\cellcolor{black!15}2 & \cellcolor{black!15} 87.9 & \cellcolor{black!15} 90.0 & \cellcolor{black!15} 59.2 & \cellcolor{black!15} 33.5 & \cellcolor{black!15} 0.0 & \cellcolor{black!15} 32.1 & \cellcolor{black!15} 3.6\\
 \multirow{2}{*}{16384} & 1  & 95.8 & 94.1 & 82.8 & 67.8 & 57.4 & 0.0 & 58.0\\
&\cellcolor{black!15}2 & \cellcolor{black!15} 87.8 & \cellcolor{black!15} 88.0 & \cellcolor{black!15} 69.9 & \cellcolor{black!15} 37.6 & \cellcolor{black!15} 0.0 & \cellcolor{black!15} 35.9 & \cellcolor{black!15} 3.1\\
 \multirow{2}{*}{32768} & 1  & 96.4 & 95.1 & 84.0 & 70.5 & 64.9 & 0.0 & 61.8\\
&\cellcolor{black!15}2 & \cellcolor{black!15} 89.5 & \cellcolor{black!15} 88.0 & \cellcolor{black!15} 73.3 & \cellcolor{black!15} 39.8 & \cellcolor{black!15} 0.0 & \cellcolor{black!15} 35.4 & \cellcolor{black!15} 4.4\\
 \multirow{2}{*}{65536} & 1  & 96.1 & 95.8 & 86.7 & 72.7 & 65.8 & 0.0 & 63.8\\
&\cellcolor{black!15}2 & \cellcolor{black!15} 89.0 & \cellcolor{black!15} 88.3 & \cellcolor{black!15} 75.1 & \cellcolor{black!15} 37.9 & \cellcolor{black!15} 0.0 & \cellcolor{black!15} 38.2 & \cellcolor{black!15} 2.0\\
\bottomrule 
\end{tabular}
 \end{table}
%
%
%
%
\begin{table}[t]
\centering
 \caption{Predictions' \emph{Correctness} after $n$ steps, with the door opened at the 2048$^{th}$ step.}
 \label{tab: Correctness Door Opened}
\setlength{\tabcolsep}{2mm}
 \renewcommand{\arraystretch}{1.1}
\fontsize{9pt}{9pt}\selectfont 
\begin{tabular}{c|r| r r r r r r r } 
\toprule\rotatebox{90}{\parbox{.8cm}{Nb of\\ Steps}} &\rotatebox{90}{Room} & \rotatebox{90}{OK}& \rotatebox{90}{KO} & \rotatebox{90}{Fail.} & \rotatebox{90}{Wall} & \rotatebox{90}{Cold}& \rotatebox{90}{Sound} & \rotatebox{90}{\parbox{.8cm}{Box \\ Name}}\\
\hline
\rule{-2pt}{2.2ex}
 \multirow{2}{*}{2048} & 1  & 92.7 & 91.4 & 89.6 & 84.9 & 90.1 & 0.0 & 82.2\\
&\cellcolor{black!15}2 & \cellcolor{black!15} 84.3 & \cellcolor{black!15} 84.8 & \cellcolor{black!15} 50.6 & \cellcolor{black!15} 31.5 & \cellcolor{black!15} 0.0 & \cellcolor{black!15} 0.0 & \cellcolor{black!15} 0.0\\
 \multirow{2}{*}{4096} & 1  & 95.3 & 92.2 & 82.5 & 76.5 & 89.5 & 0.0 & 89.8\\
&\cellcolor{black!15}2 & \cellcolor{black!15} 92.0 & \cellcolor{black!15} 87.6 & \cellcolor{black!15} 60.6 & \cellcolor{black!15} 46.6 & \cellcolor{black!15} 0.0 & \cellcolor{black!15} 52.0 & \cellcolor{black!15} 0.7\\
 \multirow{2}{*}{8192} & 1  & 95.2 & 92.5 & 86.3 & 76.3 & 97.0 & 0.0 & 90.1\\
&\cellcolor{black!15}2 & \cellcolor{black!15} 90.7 & \cellcolor{black!15} 87.7 & \cellcolor{black!15} 63.2 & \cellcolor{black!15} 41.9 & \cellcolor{black!15} 0.0 & \cellcolor{black!15} 59.2 & \cellcolor{black!15} 0.7\\
 \multirow{2}{*}{16384} & 1  & 96.2 & 93.1 & 86.3 & 78.6 & 94.9 & 0.0 & 91.6\\
&\cellcolor{black!15}2 & \cellcolor{black!15} 92.6 & \cellcolor{black!15} 89.6 & \cellcolor{black!15} 64.4 & \cellcolor{black!15} 48.0 & \cellcolor{black!15} 0.0 & \cellcolor{black!15} 71.7 & \cellcolor{black!15} 0.8\\
 \multirow{2}{*}{32768} & 1  & 96.0 & 94.1 & 88.4 & 80.2 & 97.6 & 0.0 & 92.8\\
&\cellcolor{black!15}2 & \cellcolor{black!15} 92.5 & \cellcolor{black!15} 90.9 & \cellcolor{black!15} 67.5 & \cellcolor{black!15} 50.4 & \cellcolor{black!15} 0.0 & \cellcolor{black!15} 69.9 & \cellcolor{black!15} 1.2\\
 \multirow{2}{*}{65536} & 1  & 96.4 & 95.3 & 89.4 & 81.0 & 98.2 & 0.0 & 95.1\\
&\cellcolor{black!15}2 & \cellcolor{black!15} 92.5 & \cellcolor{black!15} 91.6 & \cellcolor{black!15} 68.2 & \cellcolor{black!15} 48.2 & \cellcolor{black!15} 0.0 & \cellcolor{black!15} 68.1 & \cellcolor{black!15} 0.7\\
\bottomrule 
\end{tabular}
 \end{table}

It should be remarked that this learning did not lead to significant loss of previous knowledge regarding the first room: a moderate drop in \emph{Hit Rates} can be observed for boxes' names, \ColdFname\ and walls, but not for \OKFname, \KOFname, and \FailFname. Furthermore, \emph{Correctness} is barely impacted.
This is due to the use of selective forgetting in the learning process (remember O- and A- neurons'  boosting of inputs in Section \ref{sec: NN functioning}). As neurons encoding more particular concepts (such as detailed descriptions of particular locations) are less often reactivated, they tend to be reallocated over time, which leads to the loss of the particular features they encode. But neurons encoding more general concepts are more often reactivated, which preserves them from reallocation and forgetting.
As a result, particular and less-used concepts tend to fade out over time, while more general and well-used concepts are preserved.

\smallskip
Accordingly, the agent's ability to make appropriate decisions in the first room was not impacted by the door being opened. In fact, the bar chart of \OKFname, \KOFname, and \FailFname\ outcomes obtained in this second run of tests showed no visible difference with the one obtained with the door closed shown in Figure~\ref{fig:Pictures/OK ResultsR1R2}.A.
%


\subsection{Experiment \#3}
To better assess the agent's ability to update its knowledge to accommodate environment changes, a third experiment was conducted. 
As previously, the agent was kept in the first room for 2048 steps before the door was opened, but then it was prompted to run 50 series of 100 steps. Half way through these, the \SoundFname\ feature was moved from the boxes $(3, 3)$ - $( 8, 3)$  to the boxes $(3, -1)$ - $( 8, -1)$ (as in Figure \ref{fig: Universe}.C).  
At the end of each of these series of steps, the agent was asked for its predictions about the outcome of two actions having \SoundFname\ as the only available information about the depart location: the first one when considering a north move, and the second, when considering a south move. It was recorded whether \NWallFname\ (resp., \SWallFname) was predicted. Other predicted features, if any, were discarded. 

\begin{figure}[t]
 \includegraphics[width=\columnwidth]{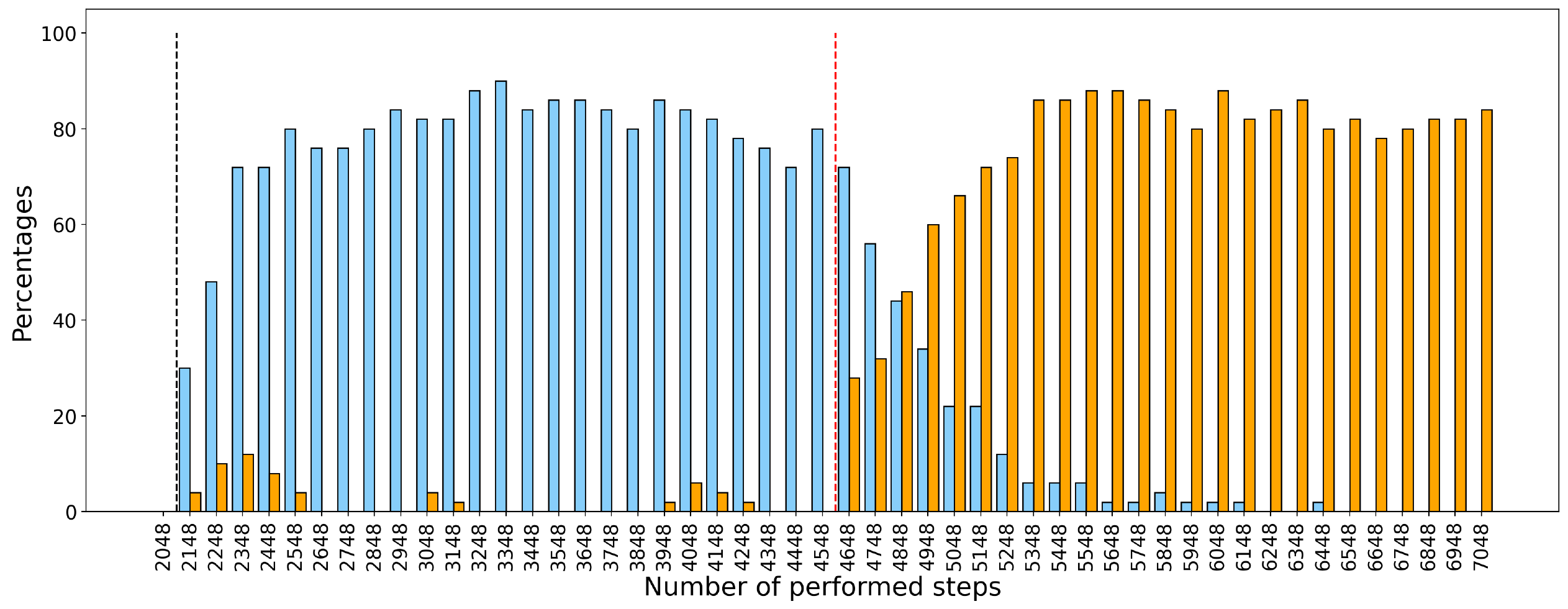}
 \caption{Prediction of walls in arrival boxes given \SoundFname\ as information about the depart location. Black dotted line: door opens; Red dotted line: sound changes location; Blue:  \NWallFname\  predictions considering a north move;  Orange: \SWallFname\  predictions considering a south move (percentages over 50 trials).}
 \label{Fig: SoundRule test}
 \end{figure}

 Figure \ref{Fig: SoundRule test} shows the percentage of  \NWallFname\ (resp., \SWallFname) predictions  after each series of steps.
These results show that after the door opened the agent first learned to predict a north wall in the arrival box when considering a north move, but that once the \SoundFname\ feature was moved to its ``down'' position it stopped doing so and learned to predict a south wall in the arrival box when considering a south move  instead.
This learning was both fast and robust, considering that the agent had very few learning occasions. Indeed less than 0.6\% (on average) of taken steps were north moves from a box with sound, and similarly for south moves from a box with sound. This means that very few experiences were enough for the agent to update its knowledge, which is consistent with the results obtained in the first experiment.

\subsection{Computing time}
Computing time was simply estimated by running trials consisting in a single series of 2024 steps. An average of 5.9 seconds (standard deviation 0.16) was obtained on a conventional computer. However this is to be taken as  an upper bound, since for practical reasons some testing carried out in the course of the series was not deactivated. No attempt to optimize the computing time was made, as it seems less critical in the case of online learning of autonomous agents which can learn while physically performing their actions.
%


\section{Conclusion and Future Developments}\label{sec: Conclusion}

In this paper a proof of concept of the architecture of a fully autonomous agent learning action laws online and accommodating environment changes was provided. 
This agent relies on general concepts to handle new situations and dynamically adjusts its concepts to its current environment. This makes it well suited for open worlds: if a new door were to open to a third room with new objects and laws, it would learn them just as it did in the second room. 
Of course, this would come at the cost of the forgetting of its least used concepts, but these are precisely the ones it needs the less. 
In fact, the agent's ability to selectively forget its least used knowledge ensures that it will always be able to adapt to new environments by replacing old unused concepts by new useful ones. 

It should be stressed that this architecture does not crucially depends on the specific type of spiking neurons used to implement it nor on the details of its implementation. 
Any similar architecture based on a spiking neural network using some STDP-like learning rules would probably work, provided that it retains the approach's main ideas. 
Notably, strict constraints on output neurons' connections weight sums associated with flexible constraints on input neurons' forward connections weight sums allow to regulate learning and keep the network balanced; 
target numbers of neurons for spiking and learning associated with dynamic learning thresholds ensure that the network will always have neurons to learn a new input; 
boosting the inputs received by the neurons having not spiked for a long time allows to preferentially select those encoding less used knowledge for learning new inputs, and to prevent in this manner catastrophic forgetting;
forcing a few neurons to learn particular concepts counterbalances STDP's tendency towards generalization and favours the representation of concepts with diverse degrees of generality; 
finally, input modulation in the querying process promotes the firing of neurons encoding the most specific information, thus helping the agent to make accurate predictions in non-monotonic contexts. 
All these methods rely on individual neurons' properties such as the weight sum of their incoming / forward connections, the time of their last spike and other similar metrics. 

\smallskip
As regards scalability to larger environments, the number of neurons needed in each layer of the network (interface, O-neurons, A-neurons) grows linearly with the number of items (features, object concepts, action concepts) to encode; the number of synapses needed on each O-neuron grows linearly with the maximum number of features to be learned for a given object/situation. 
However, it is important to observe that these are upper bounds: the agent's ability to form general concepts out of individual experiences and to use them to make decisions makes that it is not necessary to encode each individual object with all its features and each performed action.

\smallskip
Future developments of the architecture include endowing the agent with planning abilities.
To do so, a notion of \emph{executable} action law should be defined (see Section \ref{sec: Concepts and action laws}). The agent would also need to build its own set of possible situations (states) online (the set of its object concepts could probably be used to this end). 
Finally, the decision system should be augmented to represent goals and embed a cost function and a planning algorithm.
\smallskip

However further work remains to be done to allow the agent to live in more realistic environments.
Notably, it would be necessary to make the agent able to use incomplete information as input for learning and querying, as real world agents' observations are rarely complete.
 It would then be interesting to make the agent able to query its semantic memory for object properties given some partial input. 
It would also be useful to implement negation in the network to allow the agent to represent the fact that a given object does not have a given feature. Neural inhibition could probably be used for this, but the appropriate learning rules remain to be found. 
It seems that taken together these two improvements would bring the agent to draw non-monotonic inferences in the spirit of \citep{Gri16}.

It would also be suitable to make the agent able to distinguish between objects and their locations, as actions can modify one, the other, or both. 
Biological brains process the ``what'' and the ``where'' components of observations in two separate pathways before reunifying them, and this could be an inspiration source. 

A further line of research would be to investigate how an agent should decide between  \emph{Exploration} and \emph{Exploitation} modes in an open world. 
Intuitively, it seems that the choice between these two modes should depend on the agent's estimation of the risk associated with an explorative behaviour, and that the agent should refrain from entering in exploration mode in situations where the assessed risk is high, while allowing itself more exploration in situations where it is low. Yet a correct formalization of this idea remains to be found.

\bigskip
%


\subsection*{\normalsize Declaration of conflicting interests}
\vspace{-1ex}
None

\subsection*{\normalsize Funding}
\vspace{-1ex}
This work is supported by the ANR (Agence Nationale de la Recherche) project ALoRS (``Action, Logical Reasoning and Spiking networks'') [ANR-21-CE23-0018-01]

\subsection*{\normalsize Data availability}
\vspace{-1ex}
\noindent The program allowing to run and test the agent can be downloaded at \href{https://zenodo.org/records/17463567}{https://zenodo.org/records/17463567}


\bibliography{mybibfile}


\section*{Appendices}

\subsection*{1. Parameters}
The network's parameters were set as follows:

\begin{itemize}
    \item Interface neurons:
    \begin{compactitem}
        \item $\STL = 50$ (L-neurons' Spike Threshold)
        \item $\STFail\ = 40$ ($\FailureNeuron$ neuron's Spike Threshold)
    \end{compactitem}
    Lowering these spike thresholds increases the predictions' \emph{Hit Rates} while decreasing their \emph{Correctness}. Reciprocally, increasing these spike thresholds increases the predictions' \emph{Correctness} but decreases their \emph{Hit Rates}.

    \item O-neurons:
    \begin{compactitem}
        \item  $\WSO = 40$ is the number of incoming synapses on O-neurons. This value was chosen so as to allow enough redundancy between synapses to ensure robustness of learning while keeping the computational complexity as low as possible.
        \item For $o$ in \O, $\STOo$ ($o$'s forward Spike Threshold) is an integer randomly chosen in $\{22, ...,  31\}$. 
        This range of values was chosen so that about 1/2 - 3/4 of a neuron's synapses need to be activated for the neuron to reach its spike threshold.
        \item  For $o$ in \O,  $\backwardSTOo = \mathrm{floor}(\STOo/3)$.   $\backwardSTOo$  is $o$'s backward Spike Threshold. Increasing its value prevents the backward firing of O-neurons in the querying process and decreases predictions' \emph{Hit Rates}; decreasing its value decreases the predictions' \emph{Correctness}.
        \item $\boostParamO = 50$ (parameter in the boosting equation for O-neurons).  Lowering its value augments the boosting and tends to destabilize the network; increasing its value lessens the boosting, which leaves some neurons unused while other neurons are consistently reused, thus preventing the agent to retain the acquired knowledge in the long run.  
        \item $\TnbFiredO = 12$ (target number of forward firing O-neurons). Keeps the number of firing / learning O-neurons under control and prevents that too many neurons learn a same situation.
        \item  $\TnbLearningO = 6$ (target number of learning O-neurons). This parameter's value is set to half the target number of forward firing O-neurons to ensure that boosting is only used when very few neurons respond to the input (which corresponds to situations that are new to the agent).
       \item $\TnbQueryO =  6$ (target number of backward firing O-neurons in the query process). Its value is set at the target number of learning O-neurons as the neurons having learned the situation to predict are the ones that need to be reactivated in the querying process.
        \item Upper bound for $\lastSpikedO$ initialization $= 2,000$. Setting this parameter to some high value favors the use of new neurons at early stages of learning and accelerates early learning.
        \item $\noiseO = 2$ (noise threshold for O-Neurons). Noise suppression improves correctness.
    \end{compactitem}

    \item A-neurons:
    \begin{compactitem}
        \item  $\WSA = 30$ is the number of incoming synapses on each compartment of A-neurons. This value was chosen so as to allow enough redundancy between synapses to ensure robustness of learning while keeping the computational complexity as low as possible.
         \item $\learnAT = 55$ is the learning threshold for A-neurons. It ensures that only A-neurons that fit the input pair $(\inputOAi, \inputMAii)$ among those that fire are primarily selected for learning (the maximal possible input sum on the first two compartment is $2*\WSA  =  60)$.
        \item $\TnbFiredA =  \TnbLearningA =  \TnbQueryA = 4$ (target numbers of firing/learning A-neurons). 
        \item $\boostParamA = 800$ (Parameter in the boosting equation for A-neurons).  Lowering its value augments the boosting and tends to destabilize the network; increasing its value lessens the boosting, which leaves some neurons unused while other neurons are consistently reused, thus preventing the agent to retain the acquired knowledge in the long run.
         \item upper bound for $\lastSpikedA$ initialization $= 20,000$. Setting this parameter to some high value favors the use of new neurons at early stages of learning and accelerates early learning.
        \item $\noiseA = 2$ (noise threshold for A-Neurons). Noise suppression improves correctness.
        \item $\minCoefModulatedCnx = 1$, $\maxCoefModulatedCnx = 2$  (minimal and maximal coefficients for connection modulation). Favors the early spiking of A-neurons encoding more particular action concepts in the Query process.
    \end{compactitem}
\end{itemize}

\bigskip

\subsection*{2. Pseudocode}
  
%
\begin{algorithm}[H]
\caption{$\backwardSpikeL$  (backward firing of \L-neurons)}\label{backwardSpikeL}
\begin{algorithmic}[1]
\Input $(\firedO, \cnxLO)$ 

\LComment{Recall that: $\firedO \subset O$, $\cnxLO$ is a $(\card{\L} \times \card{\O})$ matrix of integers, $\noiseO$ and $STL$ are constants (respectively, the noise threshold for O-neurons and the spike threshold for L-neurons)}
\State $\firedL \gets \emptyset$ \Comment{Set of L-neurons that fired}
\State $\inhibitedLneurons \gets \emptyset$ \Comment{Set of inhibited L-neurons}
\State $\aboveT \gets \emptyset$  \Comment{dictionary with keys in \L\ and values in $\mathbb{N}$}
\For{$l \in \L$}
    \State $S = \{o \in \firedO \mid \cnxLO[l][o] >  \noiseO \}$ \Comment{Set of O-neurons with over noise threshold connections with $l$}
    \State $\backwardInputSum \gets \sum_{o \in S} \cnxLO[l][o]$
    \If{$\backwardInputSum > \STL$} 
        \State $\aboveT[l] \gets  \backwardInputSum - \STL$
    \EndIf
\EndFor
\While{$\aboveT \neq \emptyset$}
    \State $\spikingL \gets \argmaxD(\aboveT)\setminus\inhibitedLneurons$
    \State $\aboveT\ \gets \RemoveByKey(\aboveT, \argmaxD(\aboveT))$
    \For{$l \in \spikingL$}
        \State $\inhibitedLneurons\ \gets \inhibitedLneurons\ \cup~\{l' \in \L \mid \exists$ mutually exclusive features $f$ and $f'$ such that $l$ encodes $f$ and $l'$ encodes $f' \}$
    \EndFor
    \State $\firedL\ \gets  \firedL\ \cup \spikingL $ 
\EndWhile
\Statex \Return $\firedL$
\end{algorithmic}
\end{algorithm}


\begin{algorithm}[H]
\caption{\replaceAllInactiveSynapses\ (deletes and replaces all the inactive synapses of the learning neurons)}\label{replaceAllInactiveSynapses}
\begin{algorithmic}[1]

\Input $(\learningSet, \inputSet, \cnx, \inputBaseSet, \probaNewSynapses)$
\LComment{Note that: $\learningSet \subset \O$ or $\learningSet \subset \A$, depending on the case; $\inputSet \subset \inputBaseSet$ is a set of neurons; $\inputBaseSet$ is either \L, \M\ or \O, depending on the case;  $\cnx$ is a matrix of integers; $\probaNewSynapses$ is a dictionary with keys in $\inputSet$ and values in $\mathbb{R}$.}
\State $S \gets \inputBaseSet \setminus \inputSet$ \Comment{the set of inactive neurons from $\inputBaseSet$} 
\For{$n \in \learningSet$}
    \State $\decreasedWeight \gets \sum \cnx_{i \in S}[i][n]$
    \For{$i \in S$} 
        \State $\cnx[i][n] \gets 0$  \Comment{All of n's inactive synapses are deleted} 
    \EndFor
    \State $\newSynapses \gets$  \hyperref[randomChoice]{\randomChoice}$(\inputSet, \decreasedWeight, \probaNewSynapses, $ $\mathit{replace} = \mathrm{True}) $
    \Comment{vector of indices of neurons from $\inputSet$ of length $\decreasedWeight$; a same index  may occur multiple times in $\newSynapses$, each occurrence representing a new synapse from the considered input neuron.}
    \For{$i \in \newSynapses$}
        \State $\cnx[i][n]\gets \cnx[i][n] + 1$ \Comment{deleted synapses are replaced with new synapses from input neurons.}
    \EndFor
\EndFor
\Statex \Return $\cnx$

\end{algorithmic}
\end{algorithm}


\begin{algorithm}[H]
\caption{$\replaceAllSynapses$ (deletes and replaces all synapses, active or inactive, of the $\maxLearn$ neuron)}\label{replaceAllSynapses}
\begin{algorithmic}[1]

\Input $(\maxLearn, \inputLO, \cnxLO, \L, \probaNewSynapsesLO, \WSO)$
\LComment{Recall that: 
    $\maxLearn \in \learningO$, 
    $\inputLO \subset \L$, 
    $\cnxLO$ is a $(\card{\L} \times \card{\O})$ matrix of integers, 
    $\probaNewSynapsesLO$ is a dictionary with keys in $\inputLO$ and values in $\mathbb{R}$, 
    $\WSO$ is a constant (the fixed sum of connections' weights on any O-neuron).}

\For{$l \in \L$} 
    \State $\cnxLO[l][\maxLearn] \gets 0$  \Comment{delete all synapses, active or inactive,  of the considered O-neuron} 
\EndFor
\State $\newSynapses \gets$  \hyperref[randomChoice]{\randomChoice}$(\inputLO, \WSO, \probaNewSynapsesLO, $ $ \mathit{replace} = \mathrm{True})$
\Comment{vector of indices of L-neurons of length $\WSO$; a same index  may occur multiple times in $\newSynapses$, each occurrence representing a new synapse from the considered input neuron.}
\For{$i$ in $\newSynapses$}
    \State $\cnxLO[i][\maxLearn]\gets \cnxLO[i][\maxLearn] + 1$
\EndFor
\Statex \Return $\cnxLO$

\end{algorithmic}
\end{algorithm}


\begin{algorithm}[H]
\caption{\calculateInputSumsA\ (computes the sum of input received by a given compartment of A-neurons)}\label{calculateInputSumsA}
\begin{algorithmic}[1]

\Input $(\inputSet,  \cnx)$
\LComment{$\inputSet \subset \O$ or $\inputSet \subset \M$, depending on the case; $\cnx$ is matrix (of integers or real numbers, depending on the case) of dimensions $(\card{\O} \times \card{\A})$ or $(\card{\M} \times \card{\A})$, depending on the case.}
\State $\inputSum \gets \emptyset$  \Comment{dictionary with keys in $\A$ and values in $\mathbb{N}$ or $\mathbb{R}$, depending on the case.}

\For{$a \in$ \A}
    \State $\inputSum[a] \gets \sum_{n \in \inputSet} \cnx[n][a]$
\EndFor
\Statex \Return $\inputSum$
 
\end{algorithmic}
\end{algorithm}


\begin{algorithm}[H]
\caption{\sumInputsAiAii\ (adds up input sums received on A-neurons first and second compartments)}\label{sumInputsAiAii}
\begin{algorithmic}[1]

\Input $(\inputSumAi, \inputSumAii)$
\LComment{Recall that: 
    $\inputSumAi$ and $\inputSumAii$ are dictionaries with keys in \A\ and values in $\mathbb{N}$ or $\mathbb{R}$, depending on the case;
    $\noiseA$ is a constant (noise threshold for A-neurons).}
\State $\inputSumAiAii\  \gets \emptyset$  \Comment{dictionary with keys in \A\ and values in $\mathbb{N}$ or $\mathbb{R}$, depending on the case.}
\For{$a \in$ \A} 
    \If{$\inputSumAi[a] > \noiseA$  and   $\inputSumAii[a] > \noiseA$} 
        \State $\inputSumAiAii[a]\gets \inputSumAi[a]+ \inputSumAii[a]$
    \EndIf
\EndFor
\Statex \Return $\inputSumAiAii$
 
\end{algorithmic}
\end{algorithm}


\begin{algorithm}[H]
\caption{\selectLearningNeuronsA\ (selects the A-neurons that will undergo learning)}\label{selectLearningNeuronsA}
\begin{algorithmic}[1]
\Input $(\inputOAi, \inputMAii, \cnxOAi, \cnxMAii, \lastSpikedA)$
\LComment{Recall that: 
    $\inputOAi \subset \O$,
    $\inputMAii \subset \M$,
    $\cnxOAi $ and $\cnxMAii$ are matrices of integers of respective dimensions $(\card{\O} \times \card{\A})$ and $(\card{\M} \times \card{\A})$ ,
    $\lastSpikedA$ is a vector of integers of size $\card{\A}$, 
    $\learnAT$ and $\TnbLearningA$ are constants (resp. learning threshold for A-neurons and target number of learning A-neurons).}
\State $(\inputSumAiAii, \firedA)\gets$ \hyperref[spikeA]{\spikeA}$(\inputOAi, \inputMAii, \cnxOAi, \cnxMAii)$
\State $\learningA \gets \{a \in \firedA \mid \inputSumAiAii[a] > \learnAT\}$
\If{$\card{\learningA} <  \TnbLearningA$}
    \State $\boostedinputSumAiAii \gets \emptyset$  \Comment{dictionary with keys in \A\ and values in $\mathbb{R}$}
    \State $S = \{a \mid (a, v) \in inputSumAiAii\}$ \Comment{set of A-neurons that have received some input}
    \For{$a \in S\setminus\learningA$}
        \State $\boost \gets (\lastSpikedA[a]+ \boostParamA) / \boostParamA$  \LComment{$\boostParamA = 800$ is a constant; it determines the slope of the function's curve.}
        \State $\boostedinputSumAiAii[a] \gets \inputSumAiAii[a]* \boost$
    \EndFor
    \While{$\card{\learningA} <  \TnbLearningA$ and $\boostedinputSumAiAii \neq \emptyset$}
        \State $\addedNeurons \gets \argmaxD(\boostedinputSumAiAii)$ \Comment{subset of \A}
        \State $\learningA \gets \learningA\cup\addedNeurons$
        \State $\boostedinputSumAiAii \gets \RemoveByKey(\boostedinputSumAiAii, $ $ \addedNeurons)$
    \EndWhile
\EndIf
\Statex \Return $\learningA$

\end{algorithmic}
\end{algorithm}


\begin{algorithm}[H]
\caption{\calculateProbaNewSynapsesA\ (calculates O-, M- and $\FailureNeuron$ neurons' probability to establish new synapses onto A-neurons)}\label{calculateProbaNewSynapsesA}
\begin{algorithmic}[1]

\Input $(\inputSet, \cnx )$
\LComment{Note that: $\inputSet \subset \O$ or $\inputSet \subset \M$ or $\inputSet = \F$, depending on the case; $\cnx$ is a matrix of integers.}
 \State $\SGR, \probaNewSynapses \gets \emptyset$   \Comment{dictionaries with keys in $\inputSet$ and values in $\mathbb{R}$.}
\For{$n \in \inputSet$}
    \State $\fwrdCnxSum \gets \sum \cnx_{a \in \A}[n][a]$ 
    \State $\SGR[n] \gets \tanh((-\fwrdCnxSum +\mathsf{a})/\mathsf{b})+\mathsf{c}$
    \LComment{$\mathsf{a} = 400$,  $\mathsf{b} = 300$ and $\mathsf{c} = 1.2$ are constants that determine the horizontal offset, the slope and the vertical offset of the function's curve.}
\EndFor
\State $s = \sum_{n \in \inputSet}\SGR[n]$
\For{$n \in \inputSet$}
    \State $\probaNewSynapses[n]\gets \SGR[n]/s$ 
\EndFor
\Statex \Return  $\probaNewSynapses$

\end{algorithmic}
\end{algorithm}


\begin{algorithm}[H]
\caption{\calculateLR\ (calculates A-neurons' learning rates)}\label{calculateLR}
\begin{algorithmic}[1]
\Input $(\learningA, \lastSpikedA)$
\LComment{Recall that: $\learningA \subset \A$, $\lastSpikedA$ is a vector of integers of size $\card{\A}$, $\WSA$ is a constant (the sum of connections' weights on any compartment of any \A-neuron).}
\State $\LR \gets \emptyset$  \Comment{dictionary with keys in \A\ and values in $\mathbb{N}$}
\State $S \gets \{\lastSpikedA[a] \mid a \in \learningA\}$
\State $\mathit{mini}~\gets~\min(S)$
\State $\mathit{maxi}~\gets~\max(S)$
\State  $\mathit{diff}~\gets~\mathit{maxi} - \mathit{mini}$
\If{$\mathit{diff} > 0$}
    \For{$a \in \learningA$}
        \State $\LR[a]\gets~\max(1,~\min(\WSA,~\mathrm{ceil}((\lastSpikedA[a]*\WSA/\mathit{diff}) - \mathit{mini}*\WSA/\mathit{diff})))$
    \EndFor
\Else 
    \State $\randomNeuron~\gets~$\hyperref[randomChoice]{\randomChoice}$(\learningA)$
    \State $\LR[\randomNeuron]~\gets \WSA $
    \For{$a \in \learningA \setminus\{\randomNeuron\}$}
        \State $\LR[a]~\gets 1$ 
    \EndFor
\EndIf
\Statex \Return $\LR$
\end{algorithmic}
\end{algorithm}


\begin{algorithm}[H]
\caption{\replaceSomeInactiveSynapses\ (deletes and replaces some of the learning neurons inactive synapses)}\label{replaceSomeInactiveSynapses}
\begin{algorithmic}[1]

\Input $(\learningA, \inputOAiii, \cnxOAiii, \probaNewSynapsesOAiii, \LR)$ 

\LComment{Recall that: $\learningA \subset \A$; $\inputOAiii \subset \O$ or $\inputOAiii = \F$, depending on the case;  $\cnxOAiii$ is a matrix of integers of dimensions $(\card{\O \cup \F} \times \card{\A})$; $\probaNewSynapses$ is a dictionary with keys in $\inputOAiii$ and values in $\mathbb{R}$; $\LR$ is a dictionary with keys in $\learningA$ and values in $\mathbb{N}$.}
\State $S \gets \O \cup \F\setminus \inputOAiii$ 
\For{$a \in \learningA$}
    \State $\mathit{inactiveSynapsesWeightSum} \gets \sum_{n\in S}\cnxOAiii[n][a]$ 
    \State $\decreasedWeight \gets \min(\mathit{inactiveSynapsesWeightSum}, \LR[a])$ 
    \State $\mathit{inactiveSynapses} \gets ()$ \Comment{tuple of neuron indices}
    \For{$n \in S$}
        \State add  $\cnxOAiii[n][a]$ occurrences of the index ``$n$'' to $\mathit{inactiveSynapses}$
    \EndFor
    \State $\synapsesToDelete \gets$ \hyperref[randomChoice]{\randomChoice}$(\mathit{inactiveSynapses}, \decreasedWeight, \mathrm{replace=False})$
    \Comment{vector of indices of neurons from $S$ of length $\decreasedWeight$; a same neuron index  may occur multiple times in $\synapsesToDelete$, each occurrence representing a synapse to be deleted.} 
    \For{$i \in \synapsesToDelete$}
        \State  $\cnxOAiii[i][a] \gets \cnxOAiii[i][a] -1$
    \EndFor
    \State $\newSynapses \gets$  \hyperref[randomChoice]{\randomChoice}$(\inputOAiii, \decreasedWeight, \probaNewSynapsesOAiii, $ $ \mathrm{replace=True})$ 
    \Comment{vector of indices of neurons from $\inputOAiii$ of length $\decreasedWeight$; a same neuron index  may occur multiple times in $\newSynapses$, each occurrence representing a new synapse from the considered neuron to be created.}         
    \For{$i$ in $\newSynapses$}
        \State $\cnxOAiii[i][a]\gets \cnxOAiii[i][a] + 1$
    \EndFor
 \EndFor        
\Statex \Return $\cnxOAiii$

\end{algorithmic}
\end{algorithm}

\begin{algorithm}[H]
\caption{\replInactPlusSomeActSynapses\ (deletes and replaces all inactive synapses plus some active active synapses of the learning neurons)}\label{replInactPlusSomeActSynapses}
\begin{algorithmic}[1]

\Input $(\learningA, \inputSet, \cnx, \inputBaseSet, \probaNewSynapses, \LR)$
\LComment{Recall that: $\learningA \subset \A$; $\inputSet \subset \inputBaseSet$ is a set of neurons; $\inputBaseSet$ is either \O, \M\ or $\O \cup \F$, depending on the case;  $\cnx$ is a matrix of integers; $\probaNewSynapses$ is a dictionary with keys in $\inputSet$ and values in $\mathbb{R}$.}
\State $S \gets \inputBaseSet\setminus \inputSet$
\For{$a \in \learningA$}
    \State $\decreasedWeight \gets \sum_{n \in S}\cnx[n][a]$
    \For{$n \in S$} 
        \State $\cnx[n][a] \gets 0$  \Comment{Delete all inactive synapses} 
    \EndFor
    \State $\remainsToDecrease \gets \max(0, \LR[a]-  \decreasedWeight)$
    \If{ $\remainsToDecrease >0$} 
        \LComment{randomly choose $\remainsToDecrease$ active synapses and delete them}
        \State $\activeSynapses \gets ()$ \Comment{tuple of neuron indices}
        \For{$n\in \inputSet$}
            \State add $\cnx[n][a]$ occurrences of the index ``$n$'' to $\activeSynapses$
        \EndFor
        \State $\synapsesToDelete \gets$ \hyperref[randomChoice]{\randomChoice}$(\activeSynapses, ~\remainsToDecrease, ~\mathrm{replace=False})$
        \For{$i$ in $\synapsesToDelete$}
            \State  $\cnx[i][a] \gets \cnx[i][a] -1$
        \EndFor
    \EndIf
    \State $\decreasedWeight \gets \decreasedWeight +\remainsToDecrease$
    \State $\newSynapses \gets$  \hyperref[randomChoice]{\randomChoice}$(\inputSet, \decreasedWeight, \probaNewSynapses)$ 
    \Comment{vector of indices of $\inputSet$ neurons of length $\decreasedWeight$; a same neuron index may occur multiple times in $\newSynapses$, each occurrence representing a new synapse from the considered input neuron.}     
        
    \For{$i$ in $\newSynapses$}
        \State $\cnx[i][a]\gets \cnx[i][a] + 1$
    \EndFor
\EndFor 
\Statex \Return $\cnx$

\end{algorithmic}
\end{algorithm}


\begin{algorithm}[H]
\caption{\modulateCnx\ (modulates connections' efficiency for the querying process}\label{modulateCnx}
\begin{algorithmic}[1]
\Input $(\inputSet,  \cnx)$
\LComment{Note that: $\inputSet \subset \O$ or $\inputSet \subset \M$, depending on the case; $\cnx$ is a matrix of integers of dimensions $(\card{\O} \times \card{\A})$ or $(\card{\M} \times \card{\A})$, depending on the case.}
\State  $\modulatedCnx  \gets \cnx$
\State $\fwrdCnxSum \gets \emptyset$ \Comment{dictionary with keys in $\inputSet$ and values in $\mathbb{N}$} 
\For{$n \in  \inputSet$}
        \State $\fwrdCnxSum[n] \gets \sum \cnx_{a \in \A }[n][a]$
\EndFor

\State $\mathit{maxSum} \gets \max(\{\fwrdCnxSum[n] \mid n \in \inputSet\})$
\State $\mathit{minSum} \gets \min(\{\fwrdCnxSum[n] \mid n \in \inputSet\})$
\State $\mathit{diff} \gets \mathit{maxSum} - \mathit{minSum}$
\If{$\mathit{diff} > 0$}
    \State $\mathit{slope} \gets -(\maxCoefModulatedCnx - \minCoefModulatedCnx)/\mathit{diff}$
    \LComment{$\minCoefModulatedCnx\ = 1$ and $\maxCoefModulatedCnx = 2$ are constants. They set the minimal and maximal possible values of the multiplying factor $\coefMult$}
    \For{$n \in \inputSet$}
        \State $\coefMult \gets \fwrdCnxSum[n] * \mathit{slope}~+ \maxCoefModulatedCnx - \mathit{minSum} * \mathit{slope}$
        \For{$a \in$ \A}
            \State $\modulatedCnx[n][a] \gets  \mathit{Cnx}[n][a]* \coefMult $
        \EndFor
    \EndFor
\EndIf
\Statex \Return $\modulatedCnx$
 
\end{algorithmic}
\end{algorithm}


\begin{algorithm}[H]
\caption{\computeBackwardInputSumsOuF\ (computes backward input sums for O- and $\FailureNeuron$ neurons in the querying process)}\label{computeBackwardInputSumsO}
\begin{algorithmic}[1]

\Input $(\firedA, \cnxOAiii,  \backwardInputSumOuF)$
\LComment{Recall that: 
    $\firedA \subset A$; 
    $\cnxOAiii$ is a matrix of integers of dimensions \mbox{$(\card{\O \cup\F} \times \card{\A})$}; 
    $\backwardInputSumOuF$ is a vector of integers of length $\card{\O\cup\F}$;
    $\noiseA$ and $\STFail$ are constants (respectively, A-neurons' noise threshold and $\FailureNeuron$ neuron's spike threshold);
    $\backwardSTO$ is a vector of integers of length $\card{\O}$ (O-neurons' backward spikes thresholds).}
\State $\aboveBackwardSTOuF \gets \emptyset$  \Comment{dictionary with keys in $\O \cup \F\ $ and values in $\mathbb{N}$}

\For{$n \in \O \cup \F$}
    \For{$a \in \firedA$}
        \If{$\cnxOAiii[n][a] > \noiseA$}
            \State $\backwardInputSumOuF[n] \gets \backwardInputSumOuF[n] + \cnxOAiii[n][a]$
        \EndIf
    \EndFor
    \If{$n \neq \FailureNeuron$ and $\backwardInputSumOuF[n] > \backwardSTO[n]$}
        \State $\aboveBackwardSTOuF[n] \gets \backwardInputSumOuF[n]- \backwardSTO[n]$
    \EndIf
    \If{$n =  \FailureNeuron$ and $\backwardInputSumOuF[n] > \STFail$}
        \State $\aboveBackwardSTOuF[n] \gets \backwardInputSumOuF[n] -  \STFail$
    \EndIf
\EndFor
           
\Statex \Return $(\backwardInputSumOuF, \aboveBackwardSTOuF)$

\end{algorithmic}
\end{algorithm}


\begin{algorithm}[H]
\caption{\backwardSpikeO\ (backward spiking of O- and $\FailureNeuron$ neurons)}\label{backwardSpikeO}
\begin{algorithmic}[1]

\Input $(\aboveBackwardSTOuF, \inhibitedIneurons, \currentAThreshold, \backwardFiredI)$
\LComment{Recall that: $\aboveBackwardSTOuF$ is a dictionary with keys in $\O \cup \F$ and values in $\mathbb{N}$,
 $\inhibitedIneurons \subset \L \cup \F$, $\currentAThreshold \in \mathbb{R}$, and $\backwardFiredI$ is a dictionary with keys in $\L \cup \F $ and values in $\mathbb{R}$.}
\State  $\spikingO \gets \emptyset$ \Comment{Set of firing O-neurons}
\For{$n \in \argmaxD(\aboveBackwardSTOuF)$}
    \If{$n \neq \FailureNeuron$}
        \State $\spikingO \gets \spikingO \cup \{n\}$
        \If{$\FailureNeuron \notin \inhibitedIneurons$}
            \State $\inhibitedIneurons \gets \F$
        \EndIf
    \Else
        \If{$\FailureNeuron \notin \inhibitedIneurons$}
            \State $\backwardFiredI[\FailureNeuron] \gets \currentAThreshold$
            \State  $\Fail \gets~\mathrm{True}$
        \EndIf
    \EndIf
    \State $\RemoveByKey(\aboveBackwardSTOuF, \{n\})$
\EndFor
\Statex \Return $(\spikingO, \Fail, \backwardFiredI, \aboveBackwardSTOuF)$

\end{algorithmic}
\end{algorithm}


\begin{algorithm}[H]
\caption{\backwardSpikeIQuery\ (backward spiking of I-neurons in the querying process)}\label{backwardSpikeIQuery}
\begin{algorithmic}[1]
\Input $(\spikingO, \cnxLO,  \backwardInputSumL, \backwardFiredI, \inhibitedLneurons)$
\LComment{Recall that: 
    $\spikingO \subset \O$, 
    $\cnxLO$ is a matrix of integers of dimensions $(\card{\L} \times \card{\O})$, 
    $\backwardInputSumL$ is a vector of integers of length $\card{\L}$, 
    $\backwardFiredI$ is a dictionary with keys in $\L \cup \F $ and values in $\mathbb{R}$, 
    $\inhibitedLneurons \subset \L \cup \F$,
    $\noiseO$ is a constant (noise threshold for O-neurons).}
\State $\aboveSTL \gets \emptyset$  \Comment{dictionary with keys in \L\ and values in $\mathbb{R}$}
\For{$l \in \L$}
    \For{$o \in \spikingO$}
        \If{$\cnxLO[l][o] > \noiseO$} 
            \State $\backwardInputSumL[l] ~\gets \backwardInputSumL[l]~+ \cnxLO[l][o]$
        \EndIf
    \EndFor
    \If{$\backwardInputSumL[l] >  \STL$}
        \State $\aboveSTL[l]~\gets \backwardInputSumL[l]~-  \STL$
        \State $\backwardInputSumL[l] ~\gets 0$
    \EndIf
\EndFor
\While{$\aboveSTL\ \neq \emptyset$}
    \State $S \gets \argmaxD(\aboveSTL)$
    \For{$l \in S\setminus\inhibitedLneurons$}
        \If{$\nexists~c$ such that $(l, c)\in \backwardFiredI$}
            \State $\backwardFiredI(l)\gets \currentAThreshold$
            \For{$l' \in$ \L\ such that $l$ and $l'$ encode mutually exclusive features}
                \State $\inhibitedLneurons \gets \inhibitedLneurons \cup~\{l'\}$
            \EndFor
        \EndIf
    \EndFor
    \State $\aboveSTL \gets \RemoveByKey(\aboveSTL,\, S)$                                                
\EndWhile       
\Statex \Return $(\backwardFiredI, \backwardInputSumL, \inhibitedLneurons)$

\end{algorithmic}
\end{algorithm}


\begin{algorithm}[H]
\caption{\ratePredictions\ (rates possible moves for their suitability)}\label{ratePredictions}
\begin{algorithmic}[1]

\Input $\predictions$
\LComment{Recall that: \\
    $\predictions$ is a dictionary with keys in $\MotAct$ and values which are themselves dictionaries with keys in $\LF \cup \{\FailFname\}$ and values in $\mathbb{R}$.}
\State $\suitable \gets \{(\move, \confi{OK}) \mid (\OKFname, \confi{OK}) \in \predictions[\move] \}$
\State $\unsuitable \gets \{(\move, \confi{f}) \mid (f = \KOFname$ or $f = \FailFname)$  and  $(f, \, \confi{f}) \in \predictions[\move] \}$
\State $\undecided \gets \{\move \mid \nexists\conf \textrm{ such that }(\move, \conf) \in \suitable \cup \unsuitable\}$

\Statex \Return $(\suitable, \unsuitable, \undecided)$
 
\end{algorithmic}
\end{algorithm}


\begin{algorithm}[H]
\caption{\makeAChoice\ (choses between possible moves)}\label{makeAChoice}
\begin{algorithmic}[1]

\Input $(\suitable, \unsuitable, \undecided, \selectedMode)$
\LComment{Recall that: \\
    $\suitable$ and  $\unsuitable$ are set of pairs $(\move, \conf)$ where $\move \in \MotAct$ and $\conf \in \mathbb{R}$ (can be regarded as dictionaries); \\
    $\undecided \subseteq \MotAct$; \\
    $\selectedMode \in \{$``Exploration'',``Exploitation''$\}$.}

\If{$\selectedMode = $ \emph{``Exploration''} }
    \If{$\undecided\ \neq \emptyset$}
        \State $\move \gets$ \hyperref[randomChoice]{\randomChoice}$(\undecided)$
    \Else
        \State  $\mathit{decided} \gets \suitable \cup \unsuitable$ \Comment{can be regarded as a dictionary}
        \State $\move \gets$ \hyperref[randomChoice]{\randomChoice}$(\argminD(\mathit{decided}))$
    \EndIf
\ElsIf{$\selectedMode\ =$ \emph{``Exploitation'' }}
    \If{$\suitable \neq \emptyset$}
        \State $\move \gets$ \hyperref[randomChoice]{\randomChoice}$(\argmaxD(\suitable))$
    \ElsIf{$\undecided \neq \emptyset$}
        \State $\move \gets$ \hyperref[randomChoice]{\randomChoice}($\undecided$)
    \Else 
        \State $\move \gets$ \hyperref[randomChoice]{\randomChoice}$(\argminD(\unsuitable))$
    \EndIf
\EndIf
\Statex \Return $\move$
 
\end{algorithmic}
\end{algorithm}

\bigskip


\begin{algorithm}[H]
\caption{\updateLastSpikes\ (updates last spikes values)}\label{updateLastSpikes}
\begin{algorithmic}[1]

\Input $(\lastSpikesVector,\, \firedSet)$
\LComment{Recall that: \\
    $\lastSpikesVector$ is a vector of integers of length $\card{\O}$ or $\card{\A}$, depending on the case; \\
    $\firedSet \subseteq \O$ or $\firedSet \subseteq \A$, depending on the case.}
\For{$n \in \lastSpikesVector$}
    \If{$n \in \firedSet$}
       \State $\lastSpikesVector[n] \gets 1$
    \Else
        \State $\lastSpikesVector[n] \gets \lastSpikesVector[n] + 1$
    \EndIf
\EndFor
\Statex \Return $\lastSpikesVector$
\end{algorithmic}
\end{algorithm}
%


\subsection*{Computing Infrastructure}
Research was carried out on a MacBook Pro with Apple M1 Max chip (2022). Operating System: macOS-15.6-arm64-arm-64bit. We used Python with Spyder IDE.
\begin{compactitem}
\item Spyder version: 6.0.5  (standalone)
\item Python version: 3.11.11 64-bit
\item Qt version: 5.15.8
\item PyQt5 version: 5.15.9
\end{compactitem}

\end{document}